%% file: acl_latex.tex
\documentclass[11pt]{article}

\usepackage[final]{acl}

\usepackage{times}
\usepackage{latexsym}

\usepackage[T1]{fontenc}
\usepackage{kotex}

\usepackage[utf8]{inputenc}

\usepackage{microtype}

\usepackage{inconsolata}

\usepackage{graphicx}
\usepackage[table]{xcolor}
\usepackage{booktabs}
\usepackage{multicol}
\usepackage{multirow}
\usepackage{tabularx}
\usepackage{array}
\usepackage{listings}
\usepackage{tcolorbox}
\tcbuselibrary{listings}
\usepackage{subcaption}
\usepackage{xcolor}
\usepackage{amsmath}
\usepackage{amssymb}
\usepackage{enumitem}
\usepackage{fvextra}               %
\usepackage{hyperref}              %

\newcolumntype{L}[1]{>{\raggedright\arraybackslash}p{#1}}
\newcommand\blfootnote[1]{%
  \begingroup
  \renewcommand\thefootnote{}\footnote{#1}%
  \addtocounter{footnote}{-1}%
  \endgroup
}

\newcommand{\acc}{\mathrm{acc}}
\newcommand{\en}{\mathrm{EN}}
\newcommand{\tgt}{\mathrm{tgt}}

\title{From Unverifiable Domain Text to Cross-Lingual RLVR \\ via Code-Switched Task Synthesis}

\author{
  \textbf{Cheonbok Park}$^{1,2}$\textsuperscript{*} \quad
  \textbf{Jihwan Kim}$^2$\textsuperscript{*$\ddagger$} \quad
  \textbf{Shinyoung Joo}$^1$ \quad
  \textbf{Joosung Lee}$^1$ \quad \\
  \textbf{Hwiyeol Jo}$^{1}$ \quad  
  \textbf{Jeonghoon Kim}$^{1,2}$\textsuperscript{$\dagger$}\quad  
  \textbf{Jaegul Choo}$^{2}$\textsuperscript{$\dagger$} \\
  $^1$NAVER Cloud\quad$^2$KAIST AI \\
  \texttt{\{cbok.park,jeonghoon.samuel\}@navercorp.com}\\
  \texttt{\{jihvvan.kim,jchoo\}@kaist.ac.kr}
}

\begin{document}
\maketitle

\begin{abstract}

Reinforcement Learning with Verifiable Rewards (RLVR) can improve reasoning in
large language models, but specialized non-English domains expose a data
mismatch. Raw English domain text is often available, while target-language
RLVR tasks with verifiable rewards are scarce. 
We introduce \textbf{code-switched task synthesis for
cross-lingual RLVR}, a data-centric approach that derives target-language and
code-switched RLVR task variants from otherwise unverifiable English
domain text. Using fill-in-the-middle multiple-choice QA
as a seed task format, we synthesize automatically verifiable tasks
from domain passages and reshape them into cross-lingual variants. These
variants preserve the same evidence-answer relation but change how English and
target-language evidence are presented during RLVR.  On Korean medical benchmarks and cybersecurity benchmarks across four target
languages, code-switched training outperforms matched monolingual and
bilingual-mixture baselines on average target-language performance while
maintaining comparable English performance. Additional analyses provide evidence of better target-language
input understanding and stronger cross-lingual alignment in both model outputs
and internal representations. These results suggest that
code-switched training helps target-language inputs access domain reasoning
learned more strongly in English.

\end{abstract}

\blfootnote{$^*$Equal contribution.}
\blfootnote{$^\dagger$Corresponding authors.}
\blfootnote{$^\ddagger$Work done during NAVER Cloud residency program.}
\section{Introduction}
\label{sec:intro}

Reinforcement Learning with Verifiable Rewards (RLVR) has become a powerful post-training method for improving the reasoning ability of large language models~\citep{shao2024deepseekmath, guo2025deepseek}. Its effectiveness depends on tasks whose outcomes can be automatically verified and converted into reward signals. Mathematics and programming naturally provide such tasks through exact answers, symbolic checks, or execution, but the same assumption does not generally hold for knowledge-intensive domains. %
Although raw text is abundant, scaling the creation of high-quality verifiable problems is difficult and time-consuming.
We study this setting using cybersecurity and medicine as controlled domain testbeds~\citep{zhang2025medrlvr,su2025crossing}. 
This limitation is more severe outside English, where domain-specific multilingual corpora are scarce and rewardable reasoning data is even scarcer~\citep{dobler2026macereason,wang2026polymath}. Thus, a central obstacle for cross-lingual domain RLVR is not only how to obtain multilingual data, but how to turn that data into verifiable training signals.

A practical way to construct verifiable supervision from domain text is
fill-in-the-middle multiple-choice question answering (FIM-MCQA)
synthesis~\citep{lu2026golden,ren-etal-2025-alignment}. FIM-MCQA masks a span
from a domain passage, treats the removed span as the correct answer, and
rewards the model for selecting it. This turns otherwise unverifiable raw text
into an automatically verifiable task. In multilingual settings, the verified seed task
can be translated into a target language to construct target-language training
data. 

Yet translation alone may not provide the most effective transfer signal.
Multilingual reasoning gaps often stem from input understanding or
cross-lingual alignment failures~\citep{kang2025multilingual,son2026pushing},
and specialized domains may require target-language inputs to access stronger
English-side domain knowledge. Code-switching offers such a bridge by placing
English and target-language evidence in the same context, but it has mainly
been studied in pretraining, supervised fine-tuning, or inference-time
settings~\citep{chen-etal-2024-breaking,zhu-etal-2024-question,
etxaniz-etal-2024-multilingual,yoo2025csicl, lu2025syncs, yoo2025cscl}. We
instead ask how to make code-switching part of the verifiable RLVR training
signal itself.

We introduce \textbf{code-switched task synthesis for cross-lingual RLVR} as a
data-construction approach for specialized domains where English text is
available but target-language RLVR data with verifiable rewards is scarce. We
synthesize FIM-MCQA seed tasks from cybersecurity and medical corpora and derive
target-language and code-switched training variants that preserve the same
evidence-answer relation. This design keeps the training signal compatible with
standard answer-based RLVR while changing how the domain evidence is presented
across languages. We evaluate the resulting data on Korean medical and cybersecurity benchmarks and in Swahili, Hindi, and Japanese cybersecurity transfer settings. The code-switched data improves target-language performance over monolingual and bilingual-mixture baselines, with comparable English performance. Input-level, representation-level, and reasoning-trace analyses suggest that code-switched training helps models leverage English-dominant domain reasoning for target-language beyond translated and bilingual variants.

\paragraph{Contributions.}
We introduce code-switched task synthesis for cross-lingual RLVR as a data-construction approach for domains where English text is available but target-language RLVR data with verifiable rewards is scarce. Starting from English FIM-MCQA seed tasks, we derive target-language and code-switched training variants that preserve the same evidence-answer relation and remain compatible with standard answer-based RLVR. Under matched supervision, code-switched data improves target-language performance over English-only, target-language-only, and bilingual-mixture training while maintaining comparable English performance. Analyses of answer consistency, input understanding, representations, and generated reasoning suggest that code-switched training helps target-language inputs access English-side domain reasoning beyond translated and bilingual variants.

\section{Related Work}
\label{sec:related}

\subsection{Verifiable Data Construction for RLVR}

Reinforcement learning with verifiable rewards (RLVR) has become an effective
post-training paradigm for improving reasoning
~\citep{guo2025deepseek,shao2024deepseekmath}. Math and programming tasks
naturally support RLVR because their answers can often be checked directly,
for example by exact matching or execution. In specialized, data-scarce
domains, the available data often consists of raw documents rather than
rewardable MCQA examples. Building verifiable MCQA datasets by hand is costly
and time-consuming.

Recent work addresses this challenge by constructing automatically verifiable
tasks from text. General-Reasoner mines and re-verifies web QA pairs
~\citep{ma2025generalreasoner}, Nemotron-CrossThink aggregates multi-domain
reasoning corpora~\citep{akter2026nemotron}, and Golden Goose synthesizes
fill-in-the-middle multiple-choice QA (FIM-MCQA) tasks from otherwise
unverifiable text~\citep{lu2026golden,guo2026rethinkingmcq}. Domain-specific
resources also exist in cybersecurity or biomedical corpora mainly designed for supervised
training~\citep{yu-etal-2025-primus,touchent2025biomedenrichedbiomedicaldatasetenriched}. However, prior work does not
directly address how to construct verifiable domain tasks into
cross-lingual training signals when target-language data is scarce from the raw text.

Our work follows the FIM-MCQA direction but uses it for a different purpose.
A translated FIM-MCQA task gives target-language exposure, but it still keeps
English and the target language as separate versions of the same content. We
instead construct code-switched FIM-MCQA tasks, where both languages appear
within a single rewardable instance. This lets the model learn the
cross-lingual bridge through the same answer reward.

\subsection{Multilingual Reasoning Gaps and Code-Switched Transfer}

Large language reasoning models exhibit substantial gaps between English
and other languages, especially in low-resource
settings~\citep{wang2026polymath,dobler2026macereason,luo-etal-2025-mmath,ko-etal-2025-understand,shi2022language}.
Recent analyses attribute much of this gap to input understanding, where the
model fails to map non-English inputs into its dominant reasoning
language~\citep{kang2025multilingual}. This view is consistent with work
showing that hub-language reasoning often remains more accurate than forced
target-language reasoning~\citep{tam2025language,qi-etal-2025-models}, and
with mechanistic analyses identifying dominant languages as privileged
reasoning channels in multilingual LLMs~\citep{multi-lingual-logit-lens,Beyond-English-Centric-LLMs,wang-etal-2025-language-mixing}.
Taken together, these findings suggest that a central bottleneck is how
target-language inputs can access English-side domain knowledge and reasoning
behavior.

Prior work has addressed this bottleneck through both algorithmic and
data-centric routes. Algorithmic approaches modify the training framework,
reasoning language, or reward design, especially in math reasoning and machine
translation settings where automatic rewards are easier to
define~\citep{zhang2025thinknatively,feng2025mtr1zero}. Data-centric
approaches instead change the supervision or prompting interface used for
transfer, including translated or question-aligned
SFT~\citep{chen-etal-2024-breaking,zhu-etal-2024-question}, cross-lingual
alignment or bridging methods~\citep{she-etal-2024-mapo,yoon-etal-2024-langbridge},
and inference-time translation~\citep{etxaniz-etal-2024-multilingual}. These
methods show that language form can strongly affect cross-lingual reasoning,
but they do not study how mixed-language data should be synthesized for
RLVR training.

Code-switching, alternating between languages within a single
context~\citep{auer1998codeswitching}, provides a natural data-centric
interface for this problem because it places source-language and
target-language evidence in the same instance. Prior work uses code-switching
as a cross-lingual transfer signal in pretraining with synthetic
data~\citep{lu2025syncs}, supervised fine-tuning with curriculum learning or
chain-of-thought supervision~\citep{yoo2025cscl,son2026pushing}, and
inference-time demonstrations~\citep{yoo2025csicl}. We extend this line to
multilingual RLVR by synthesizing code-switched FIM-MCQA tasks from English domain corpora. The resulting tasks expose the model to English and
target-language evidence within the same rewardable instance, so cross-lingual
transfer is learned from verifiable multilingual training data rather than from
translation-only supervision. To our knowledge, this is the first data-centric
approach that uses code-switched RLVR tasks to induce cross-lingual transfer in
multilingual settings.

\begin{figure*}[t]
  \centering
  \includegraphics[angle=270, width=\textwidth]{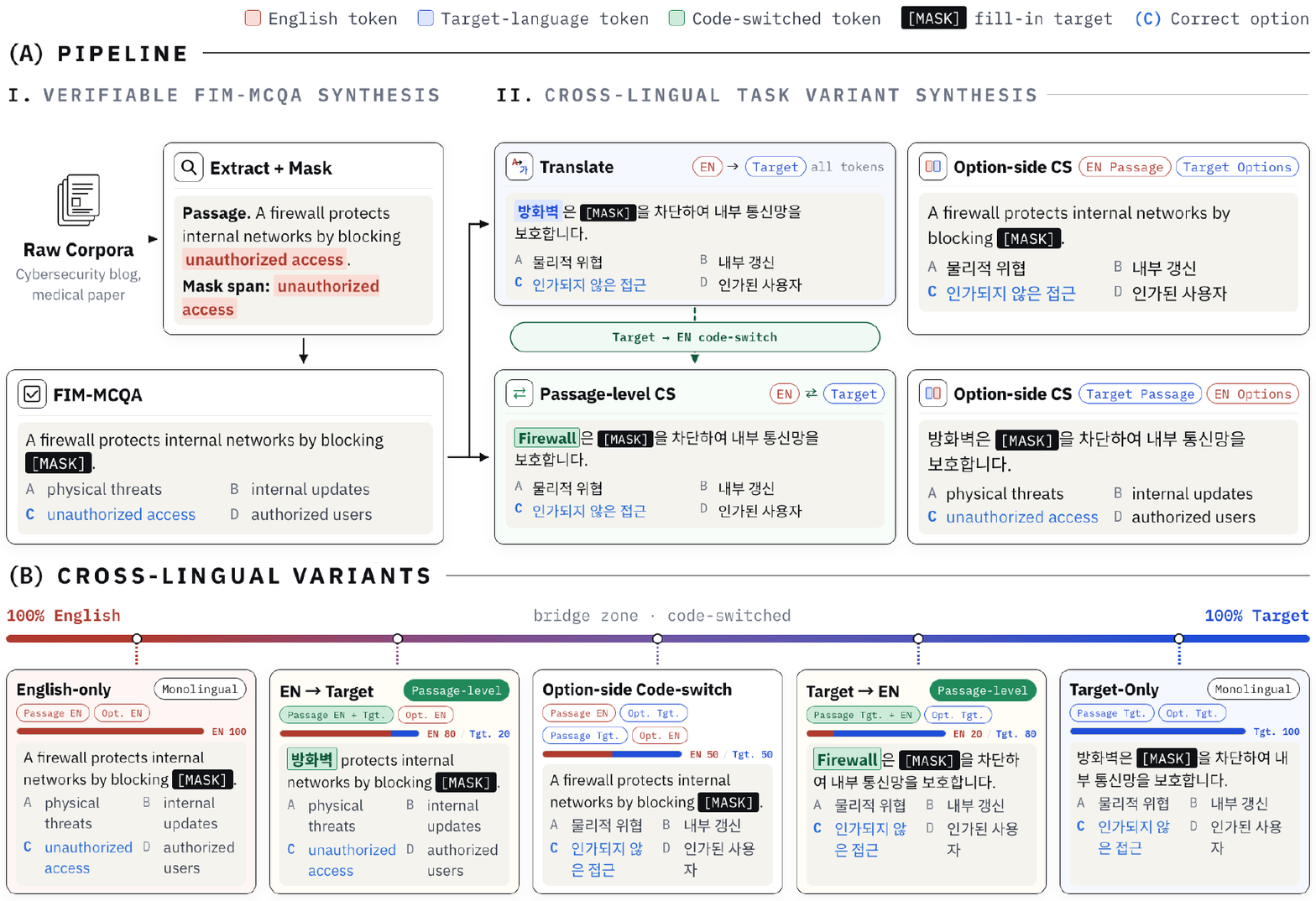}
  \caption{Data construction pipeline for cross-lingual RLVR from raw English
domain corpora. (a) We first extract educational passages from domain
raw corpora, identify answer spans, and generate distractors to synthesize
English FIM-MCQA seed tasks. We then translate each seed task into the target
language and construct code-switched variants: option-side code-switching,
which pairs passages and answer options across languages, and passage-level
code-switching, which mixes English and target-language evidence within the
passage in both directions. (b) The resulting training variants used for
controlled RLVR comparisons.}
  \label{fig:pipeline}
\end{figure*}

\section{Method}
\label{sec:method}

\subsection{Overview}
\label{sec:method-overview}

We convert raw English domain corpora into cross-lingual RLVR training tasks
for a resource-mismatch setting in which English domain text is available but
target-language verifier-ready RLVR data is scarce. We build a source pool from
cybersecurity~\citep{yu-etal-2025-primus} and medical
corpora~\citep{touchent2025biomedenrichedbiomedicaldatasetenriched}. We remove
boilerplate, non-educational pages, and low-quality documents.
Appendix~\ref{app:source} reports the filtering rules.

Figure~\ref{fig:pipeline} illustrates the pipeline. First, we synthesize English
FIM-MCQA seed tasks from the filtered documents. Each task masks a verbatim span
in a self-contained passage and uses the removed span as the answer. Second, we
derive target-language and code-switched variants from each seed task. The
variants keep the same domain content but change its language form, allowing us
to compare translation-only exposure, option-side code-switching, and
passage-level code-switching under matched supervision.

\subsection{Verifiable FIM-MCQA Synthesis}
\label{sec:method-synthesis}

For each retained source document $S$, we synthesize an English FIM-MCQA seed
task
$q^{\mathrm{en}}=(p^{\mathrm{en}}_{\mathrm{mask}},
\mathcal{A}^{\mathrm{en}}, y)$, where
$p^{\mathrm{en}}_{\mathrm{mask}}$ is a passage with one span replaced by
\texttt{[MASK]}, $\mathcal{A}^{\mathrm{en}}$ is the answer-option set, and
$y$ is the correct option index. The central constraint is that the correct
answer must originate from a verbatim span in the passage. This keeps the task
compatible with an automatic answer-option verifier during RLVR.

Rather than asking an LLM to generate the full task in one call, we decompose
synthesis into extraction, span verification, and distractor generation. The
synthesizer first extracts a self-contained educational passage $S'$ from the
filtered source document. It then selects a consecutive answer span
$a^\star$ from $S'$ and is instructed to copy the span verbatim. A deterministic
verifier checks that $a^\star$ appears in $S'$ and replaces that occurrence
with \texttt{[MASK]} to form $p^{\mathrm{en}}_{\mathrm{mask}}$. Examples that
fail this substring check are retried or discarded.

Finally, the model generates plausible distractors conditioned on $(p^{\mathrm{en}}_{\mathrm{mask}}, S', a^\star)$. The distractors are required
to match the length, format, and semantic type of the answer while avoiding
direct answer leakage. The correct span and distractors are shuffled to form
$\mathcal{A}^{\mathrm{en}}$, with the correct index recorded as $y$. This
decomposition prevents paraphrased or non-substring answers from entering the
pool while keeping the reward function simple and fully automatic. The complete 
multi-stage extraction pipeline, and prompt 
templates are provided in Appendix~\ref{app:extraction}.

\subsection{Cross-Lingual Task Variant Synthesis}
\label{sec:method-cs}

Given an English FIM-MCQA seed task,
we construct target-language and code-switched variants by changing how the same domain
content is presented across languages. We first translate the masked passage
and answer options into the target language, yielding
$q^{\mathrm{tgt}}=(p^{\mathrm{tgt}},\mathcal{A}^{\mathrm{tgt}},y)$, with the
\texttt{[MASK]} marker preserved.

Translation gives target-language exposure, but it keeps English and the target
language in separate examples. We therefore construct code-switched variants
that bring English and target-language evidence into the training task, making
language form the main intervention.

Let $\mathcal{L}=\{\mathrm{en},\mathrm{tgt}\}$, and write $p^L$ for the masked
passage in language $L$. For $u,v\in\mathcal{L}$ and $u\neq v$, we use two
code-switched variants that differ in where the language boundary is placed.

\paragraph{Option-side code-switching.}
We pair a passage in one language with answer options in the other:
\[
q_{\mathrm{opt}}^{u,v}=(p^u,\mathcal{A}^v,y).
\]
This trains cross-lingual matching between passage evidence and answer choices.

\paragraph{Passage-level code-switching.}
We mix languages inside the passage while keeping the options in the base
language:
\[
q_{\mathrm{pass}}^{u\to v}
=
(c_{u\to v}(p^u),\mathcal{A}^u,y).
\]
Here, $c_{u\to v}$ replaces selected content-bearing spans in $p^u$ with fluent
translations in language $v$.

We generate both directions, $\mathrm{en}\!\to\!\mathrm{tgt}$ and
$\mathrm{tgt}\!\to\!\mathrm{en}$. The first introduces target-language spans
into English domain passages, while the second starts from target-language
passages and inserts English spans. This design targets cross-lingual transfer:
it trains target-language inputs to connect with English-dominant domain
evidence.

Generated passages are filtered for \texttt{[MASK]} preservation, target-language coverage, script validity, and malformed switching patterns. The exact prompt templates, generation configurations, and detailed filtering rules are reported in Appendix~\ref{app:cs-detail}.

\section{Experiments}
\label{sec:experiments}

\subsection{Experimental Setting}

\paragraph{Experimental design.}
We design the experiments to test whether code-switched RLVR improves
cross-lingual transfer in specialized domains where English resources are
abundant but target-language verifier-ready RLVR data is scarce. We use
medicine and cybersecurity as two controlled domain settings. Korean is our
main target language because it provides native medical benchmarks while
remaining linguistically distant from English. We additionally evaluate
Swahili, Hindi, and Japanese in the cybersecurity domain to broaden language
and script coverage. English serves as the source language for
FIM-MCQA synthesis and for constructing target-language and code-switched views.

\paragraph{Base models.}
We use three 3B-scale open-source language models:
Llama 3.2-3B-Instruct~\citep{grattafiori2024llama},
Qwen2.5-3B-Instruct~\citep{qwen2.5}, and
SmolLM3-3B~\citep{bakouch2025smollm3}.
These models provide a suitable base model for studying cross-lingual RLVR. These models can process target-language inputs, but their domain-level reasoning behavior is still largely English-dominant.
This setting allows us to test whether target-language inputs can better leverage reasoning behavior learned most strongly in English through various settings.

\input{latex/tables/medical_main}

\paragraph{Training variants and RL setup.}

We compare the original checkpoint (\textsc{Base}) with several RL-trained
variants. \textsc{RLVR-Base} is trained only on an auxiliary general-reasoning
RLVR mixture, while the domain variants additionally use the synthesized
domain FIM-MCQA pool from Appendix~\ref{app:final_data_statistics}. These
domain variants differ only in how each domain example is realized across
languages: \textsc{En-only} uses English examples, \textsc{Tgt-only} uses
target-language examples, \textsc{Bi-Mono} mixes English and target-language
monolingual views, and \textsc{Ours} combines monolingual, Option-side, and passage-level code-switched datasets while preserving the same answer
index and verifier.

We mix domain FIM-MCQA data with the auxiliary RLVR mixture to prevent the
model from overfitting to the fill-in-the-middle MCQA format and to mitigate
format collapse during RLVR training. The auxiliary mixture contains general
reasoning data, such as scientific reasoning, math, coding, and
instruction-following prompts~\citep{akter2026nemotron,deepscaler2025,
bercovich2025llamanemotronefficientreasoningmodels}, which acts as a
regularizer for retaining broad free-form reasoning ability while the model
acquires cross-lingual, domain-specific reasoning skills. Empirically, we found
this mixing strategy to yield more stable RLVR training than using the
domain-specific FIM-MCQA pool alone.

All RL variants are trained with DAPO~\citep{yu2025dapo}. We sample $16$
rollouts per prompt, use a learning rate of $1\times10^{-6}$, and train for two epochs. Each domain variant uses the same auxiliary sampling
ratio. Additional details on the variants and training configurations are
provided in Appendix~\ref{app:appendix-training-details}.

\input{latex/tables/cyber_main}

\paragraph{Evaluation.}
We evaluate target-language and English performance in both domains. For medicine, we use MedQA~\citep{jin2020disease}, MedMCQA~\citep{pmlr-v174-pal22a}, KorMedMCQA-Doctor~\citep{kweon2024kormedmcqa}, and KMMLU-Pro-Medical~\citep{hong2025kmmlureduxkmmluproprofessionalkorean}. For cybersecurity, we use CTI-MCQ, CyberMetric, and SecEval~\citep{alam2024ctibench, tihanyi2024cybermetric,li2023seceval}. We evaluate target-language versions of the cybersecurity benchmarks in Korean, Swahili, Hindi, and Japanese. Because this task primarily targets non-English performance, we select the best checkpoint based on target-language validation scores. The translation process and subsequent quality evaluations of translated benchmarks are described in Appendices~\ref{app:eval_data_translation_details} and~\ref{app:evaluation_dataset_translation_quality_evlaution}, respectively.

\subsection{Experimental Results}
Table~\ref{tab:transposed-domain-ablation-v3} shows the medical results for Qwen2.5-3B-Instruct and Llama-3.2-3B-Instruct. RLVR improves performance across languages, regardless of whether the domain examples are presented in English, Korean, or mixed-language form. The gains are particularly clear when the training mixture includes in-domain medical FIM-MCQA data. \textsc{En-only} and \textsc{Tgt-only} often outperform \textsc{RLVR-Base}, showing that the improvement comes not only from additional RL optimization but also from rewardable examples in the target domain.

We next compare different language realizations of the same domain data. \textsc{Bi-Mono} exposes the model to both English and Korean, but it does not consistently outperform the monolingual domain baselines. This suggests that simply adding translated views or mixing monolingual examples across languages is not enough to reliably improve target-language reasoning.

The main improvement appears when code-switched examples are added. Across both medical model families, \textsc{Ours} gives more consistent gains on Korean benchmarks than the monolingual and bilingual baselines. On Korean-native benchmarks such as KMMLU-Pro-Medical and KorMedMCQA-Doctor, it improves over all non-code-switched baselines by more than one point in several settings. It also improves translated medical QA performance over both \textsc{En-only} and \textsc{Tgt-only}, indicating that code-switching helps not only on native Korean benchmarks but also on Korean versions of English medical reasoning tasks. 

A similar pattern appears beyond this Korean medical setting. As presented in Table~\ref{tab:main-domain-ko-sw_v2}, cybersecurity results across four target languages show that code-switched RLVR again yields stronger average target-language performance than monolingual or bilingual-mixture training. English performance, reported as \textit{Avg (En)}, remains comparable across variants. These results broaden the language coverage of the main finding, although the size of the improvement varies across individual benchmarks, models, and domains. Overall, the results support code-switched training as an effective cross-lingual transfer signal that helps the model apply domain reasoning learned more strongly in English to target-language inputs. Full results, including base-model performance and per-benchmark English scores, are provided in Appendix~\ref{app:full_evaluation_results}. We also report the multi-seed experiment on the Korean cybersecurity setting in Appendix~\ref{app:multiseed}.

\subsection{Component Ablation}
\label{sec:component-ablation}

To examine how each form of code-switching contributes to the
target-language improvement, we conduct a component ablation in the Korean
cybersecurity setting with SmolLM3-3B. All ablation variants follow the same
experimental setup and differ only in the included code-switching component.
\textsc{Opt-only} augments \textsc{Bi-Mono} with option-side code-switching,
\textsc{Pass-only} augments it with passage-level code-switching, and
\textsc{Ours} combines both components.

Table~\ref{tab:component-ablation-compact} shows that both code-switching
components improve the target-language average over \textsc{Bi-Mono}.
Passage-level switching yields a larger improvement than option-side
switching, while combining both components produces the strongest
target-language average and outperforms either component alone on all three
target-language benchmarks. English performance remains comparable across the
variants.  These results suggest that passage-level switching contributes more
strongly in this setting, while option-side switching provides an additional
benefit when combined with passage-level switching. Detailed results are provided in Appendix~\ref{app:ablation}.

\input{latex/tables/component_ablation_compact}

\subsection{Cross-lingual Consistency}
\label{sec:analysis-consistency}

To test whether code-switched RLVR reduces the language barrier between
English and Korean inputs, we measure accuracy-preserving cross-lingual
consistency on parallel cybersecurity questions. For each English--Korean
question pair, we partition model predictions into three mutually exclusive
cases: both languages correct, English-only correct and Korean-only correct. The both-correct rate captures language-invariant correct decisions,
whereas the English-only and Korean-only gap captures directional answer asymmetry.

As shown in Table~\ref{tab:xling_consistency},
\textsc{Ours} achieves the strongest accuracy-preserving consistency
and reduces asymmetry. This suggests that the code-switched data
does more than add translated supervision: it helps Korean input elicit the
same answer decisions as their English counterparts, providing empirical
evidence for the cross-lingual bridge.

\section{Analysis}
\label{sec:analysis}
Table~\ref{tab:xling_consistency} shows that  Code-switched RLVR improves cross-lingual consistency, but the source of this
gain is not obvious from answer accuracy alone. Following the view that
multilingual reasoning gaps arise when target-language inputs fail to reach the
model's dominant, often English-proximal, reasoning pathway, we test whether \textsc{Ours} reduces this gap at
three levels: input understanding, internal representation, and generated
reasoning behavior.

\input{latex/tables/consistency_table_main}

\subsection{In-domain Translation Quality}
\label{sec:analysis-understanding}

We first test whether the \textsc{Ours} gain begins at the input-understanding stage.
Prior work argues that multilingual reasoning gaps often arise because the
model fails to map non-English inputs into the dominant language of its
reasoning trace, typically English~\citep{kang2025multilingual}. We therefore
use in-domain Korean-to-English translation quality as a diagnostic proxy for
how well a model recovers the English-side semantic content of Korean
questions. We compare this score with the reasoning performance ratio, defined
as Korean accuracy divided by accuracy on the matched English benchmark.
Details of the translation prompt, GEMBA-DA scoring, and metrics are
provided in Appendix~\ref{app:gemba-da}.

\begin{figure}[t]
\centering
\includegraphics[width=\linewidth]{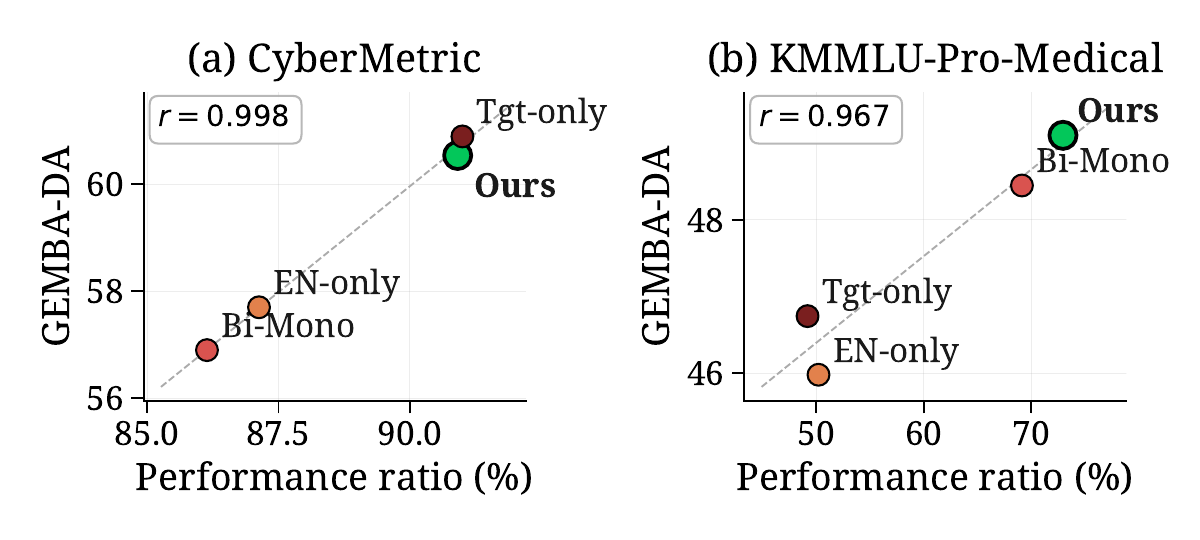}
\caption{Reasoning performance ratio vs. in-domain Korean-to-English translation quality for (a) CyberMetric (SmolLM3-3B) and (b) KMMLU-Pro-Medical (Llama-3.2-3B-Instruct). Dashed lines show linear trends; $r$ denotes Pearson correlation.}
\label{fig:ratio_indomain}
\end{figure}

Figure~\ref{fig:ratio_indomain} shows that in-domain translation quality
closely tracks retained Korean reasoning performance. The correlation is
near-perfect on CyberMetric ($r=0.998$) and remains high on
KMMLU-Pro-Medical ($r=0.967$). In both domains, the code-switched model lies
toward the upper-right region, indicating that it both recovers the
English-side meaning of Korean inputs more accurately and preserves a larger
fraction of its English reasoning performance. This supports the view that
code-switched RLVR reduces the target-language understanding bottleneck before
reasoning begins.

\subsection{Latent Reasoning State Alignment}
\label{sec:repr}

The translation analysis suggests that part of the \textsc{Ours} gain comes
from better semantic recovery of Korean inputs. We next ask whether this gain
is visible at the point where the model begins to reason. We analyze the
SmolLM3-3B ablation models on a fixed parallel English--Korean subset of
CyberMetric. For each question pair, we append the same opening reasoning
token, \texttt{\textless think\textgreater}, and extract the hidden state at
that token before any reasoning chain-of-thoughts(CoTs) are generated. We measure
English--Korean alignment across layers with Mean-Centered Cosine and Average
Neuronwise Correlation (ANC)~\citep{del-fishel-2022-cross}. Details of experiment settings and metric definitions are provided in
Appendix~\ref{app:repr-details}.

\begin{figure}[t]
\centering
\includegraphics[width=\linewidth]{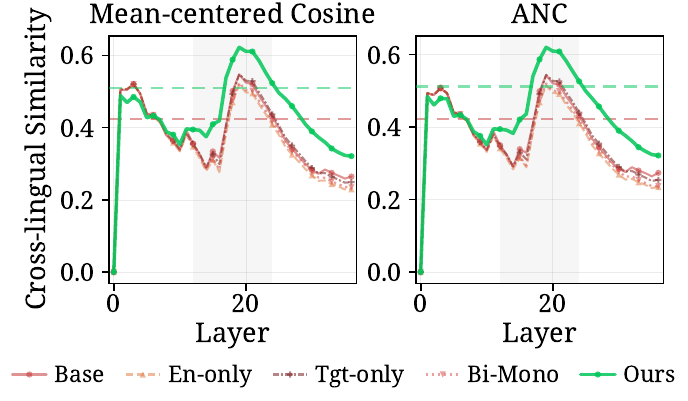}
\caption{Layer-wise English--Korean representation alignment on CyberMetric,
measured at the opening \texttt{\textless think\textgreater} marker.
Our method shows stronger middle-to-upper layer alignment than
the non-code-switched baselines. The gray area marks middle layers 12--24.}
\label{fig:repr}
\end{figure}

Figure~\ref{fig:repr} shows a common depth pattern across models:
English--Korean similarity rises in the middle layers, where cross-lingual
semantic representations are expected to become more comparable. The important
difference is the magnitude of this convergence. In the highlighted layer band,
\textsc{Ours} consistently achieves higher similarity than the non-code-switched
variants across both measures. Thus, the baselines do show some
middle-layer alignment, but their English and Korean reasoning-entry states
remain less tightly matched than those of \textsc{Ours}. This is especially
informative because \textsc{Bi-Mono} already includes
bilingual supervision; exposure to both languages alone does not produce the
same level of alignment as adding code-switched training dataset.

This depth profile is consistent with prior findings that multilingual models
often form their most comparable cross-lingual semantic states in the middle
layers \citep{muller2021first,liu2025middle}, and with the view that
non-English inputs may pass through an English-proximal reasoning space before
target-language realization \citep{multi-lingual-logit-lens}. Under this
interpretation, code-switched RLVR does not create cross-lingual alignment from
scratch. Instead, it amplifies the natural middle-layer convergence and
narrows the English--Korean gap at the reasoning-entry state.

\subsection{Generated Reasoning Trace Alignment}

The representation analysis above suggests that code-switched RLVR better aligns English and target-language inputs. We further examine whether this alignment is reflected in the model's generated reasoning CoTs. To this end, we conduct a pairwise LLM-as-a-judge evaluation that compares \textsc{Ours} with monolingual RLVR baselines. For each example, the judge is given English and target-language outputs from two systems and selects the system whose two outputs are more semantically consistent in high-level reasoning, problem decomposition, and solution strategy. We use GPT-5.5 as the judge model and swap the order to mitigate position bias. The full evaluation details are provided in Appendix~\ref{app:appendix_judge}.

Table~\ref{tab:judge_main} shows that \textsc{Ours} consistently outperforms \textsc{En-only} across all evaluated model-domain settings, with gains of $+8.3$, $+8.2$, and $+4.6$ points for Qwen2.5-3B-Instruct, Llama-3.2-3B-Instruct, and SmolLM3-3B, respectively. Compared with \textsc{Tgt-only}, \textsc{Ours} is comparable in the medical domain and substantially better in cybersecurity. These results suggest that code-switched RLVR helps the model leverage English-side domain reasoning when solving target-language problems.

\input{latex/tables/pairwise_judge_main}

\section{Conclusion}
This work proposes  cross-lingual RLVR in specialized domains as a data construction
problem to alleviate data scarcity issue in the RLVR framework. 
We build on FIM-MCQA synthesis to obtain rewardable seed tasks from English cybersecurity and medical
documents, then reshape those tasks into target-language and code-switched
training data. This lets us study how the language form of the same domain
evidence affects transfer under a standard RLVR training. 
The results show that our synthesized code-switched data provides stronger
target-language transfer than translation-only or bilingual-mixture training,
while English performance remains comparable. The analyses give a consistent
explanation. Code-switched training improves answer agreement across languages,
supports better recovery of target-language input meaning, strengthens
English--target representation alignment, and produces more consistent
reasoning traces. These findings suggest that code-switched task synthesis is a
practical way to help models leverage English-dominant domain reasoning for
target-language inputs in data-scarce multilingual domains.

\section*{Limitations}

Our study uses a controlled but computationally demanding RLVR setup. We train
multiple domain--language--model variants with the same RL algorithm and data
mixture, and each training run requires approximately $2{,}500$ H100 GPU
hours. This cost prevents exhaustive search over training schedules, data
mixtures, rollout budgets, and model scale. Prior work suggests that longer or
more carefully staged RL training~\citep{liu2026prorl}, as well as larger
rollout budgets~\citep{hu2026brorl}, can further affect reasoning performance. We
therefore interpret our results as evidence for the effectiveness of
code-switched task synthesis under a practical compute budget, rather than as a
fully optimized upper bound. For the same reason, we focus on 3B-scale
open-source models, which make controlled comparisons feasible across domains,
languages, and training variants.

A second limitation is evaluation coverage. While English domain benchmarks are
widely available, comparable target-language benchmarks are scarce, especially
outside high-resource languages. As a result, our cybersecurity evaluation
relies partly on translated benchmarks. These translations may introduce
artifacts or differ from naturally written target-language questions. We
mitigate this issue with domain-aware translation prompts and translation
quality checks in Appendix~\ref{app:evaluation_dataset_translation_quality_evlaution}.
We also include native Korean medical benchmarks, which provide a complementary
test beyond translated evaluation.

Finally, our study covers a limited range of languages, domains, and
code-switching designs. Korean is evaluated most extensively, while Swahili,
Hindi, and Japanese are examined only in cybersecurity under narrower
experimental coverage. Prior work shows that cross-lingual transfer can vary
with the form and mixture of code-switching
data~\citep{wang-etal-2025-language-mixing}. Although our component ablation compares
option-side and passage-level switching, broader language and domain settings
and alternative switching configurations remain for future work.

\section*{Acknowledgments}
We would like to thank the anonymous reviewers for their constructive feedback and suggestions. We would like to thank the anonymous reviewers for their constructive feedback and suggestions. We extend our gratitude to Gichang Lee and Nako Sung for their unwavering support. This work was partly supported by Institute for Information \& communications Technology Planning \& Evaluation (IITP) grant funded by the Korea government (MSIT) (RS-2019-II190075, Artificial Intelligence Graduate School Program (KAIST)) and by the National Research Foundation of Korea (NRF) grant funded by the Korea government (MSIT) (No. RS-2025-00555621).

\bibliography{custom}

\newpage
\appendix
\section*{Appendix}
We provide detailed supplementary materials organized as follows:

\begin{itemize}
\item \textbf{Appendix}~\ref{app:data} describes how the training data is constructed.
  \begin{itemize}
    \item \ref{app:source}: source corpora and the data pre-filtering process.
    \item \ref{app:extraction}: verifiable FIM-MCQA synthesis.
    \item \ref{app:cs-detail}: cross-lingual task variant synthesis.
    \item \ref{app:final_data_statistics}: final data statistics.
    \item \ref{app:qualitative_examples}: qualitative examples of the processed data.
  \end{itemize}
 
\item \textbf{Appendix}~\ref{app:setup} describes the experimental setup.
  \begin{itemize}
    \item \ref{app:appendix-training-details}: training details.
    \item \ref{app:eval_data_translation_details}: evaluation dataset translation.
    \item \ref{app:evaluation_dataset_translation_quality_evlaution}: translation quality assessment of the evaluation dataset.
  \end{itemize}
 
\item \textbf{Appendix}~\ref{app:results} reports additional quantitative results.
  \begin{itemize}
    \item \ref{app:full_evaluation_results}: full evaluation results, including the no-RLVR settings.
    \item \ref{app:multiseed}: multi-seed robustness results.
    \item \ref{app:ablation}: per-benchmark component ablation results.
  \end{itemize}
 
\item \textbf{Appendix}~\ref{app:analysis} details our analyses.
  \begin{itemize}
    \item \ref{app:xling_consistency}: cross-lingual consistency comparison.
    \item \ref{app:gemba-da}: in-domain translation experiment.
    \item \ref{app:repr-details}: representation alignment analysis.
    \item \ref{app:appendix_judge}: generated reasoning trace alignment experiment.
  \end{itemize}

\item \textbf{Appendix}~\ref{app:prompts} lists the prompts utilized in our experiments.
\end{itemize}

\hrulefill
\vspace{1em}

\section{Data Construction}
\label{app:data}

\subsection{Source Corpora and Pre-filtering Details}
\label{app:source}

\paragraph{Cybersecurity}
For the cybersecurity domain, we use the Primus dataset \citep{yu-etal-2025-primus}, an open-source corpus released by Trend Micro for training cybersecurity large language models. It consists of two sub-corpora: \emph{Primus-Seed}, a manually curated, high-quality collection of documents gathered from MITRE, Wikipedia, major security vendor websites, and expert-collected Cyber Threat Intelligence (CTI); and \emph{Primus-FineWeb}, which filters cybersecurity-related texts from FineWeb using a TinyBERT classifier trained with Primus-Seed as positive samples.

We combine these two sub-corpora into Primus-Pretraining and apply an additional educational quality filter. We include only documents that score $\geq 2.2$ (on a scale of 0 to 5) using nemocurator-fineweb-nemotron-4-edu-classifier\footnote{\url{https://huggingface.co/nvidia/nemocurator-fineweb-nemotron-4-edu-classifier}}~\citep{su2025nemotron}. Although the default high-quality threshold for this classifier is 2.5, cybersecurity domain texts tend to receive slightly lower average scores compared to general-purpose educational texts. Therefore, we adopt a slightly more permissive threshold of 2.2. Combining these two sources yields a total of 500k seed documents.

\paragraph{Medical}
For the medical domain, we utilize the Biomed-Enriched dataset \citep{touchent2025biomedenrichedbiomedicaldatasetenriched}, which annotates the PubMed Open Access (PMC-OA) corpus with document types, domains, and educational quality scores. We use the commercial split of this dataset and filter the seed documents based on the following criteria: an educational score of 4 or higher, domains categorized as either clinical or biomedical, and document types restricted to reviews and clinical cases. Furthermore, we exclude the "reference" sections and require all texts to have a minimum length of 200 characters. This filtering process results in a total of 450k seed documents.

\subsection{Details of FIM-MCQA Synthesis}
\label{app:extraction}
 
Our data construction pipeline converts raw documents into verifiable training tasks through a multi-stage process. 

In the first phase, we utilize the gpt-oss-120b~\citep{agarwal2025gpt}, configured with a medium reasoning effort, to simultaneously extract an educational passage and identify the target span for masking. If the document lacks educational content, the model is instructed to generate an empty response, ensuring this data is filtered out and does not proceed to the next phase.

Next, a deterministic programmatic verifier processes the passage and the identified span $(S', t)$ to construct the masked text. This programmatic layer applies a 4-tier matching protocol, consisting of exact, strip, NFKC, and fuzzy matching, to locate the span within the passage. It then replaces the verified substring with a \texttt{[MASK]} marker to produce the masked passage $S_{\text{mask}}$. If the substring check fails across all formatting tiers, the instance is rejected.

Finally, the gpt-oss-120b~\citep{agarwal2025gpt} (again with medium reasoning effort) generates the multiple-choice distractors. The model is prompted to create distractors such that each question contains a total of nine options, including the correct answer.

To ensure the quality of the constructed dataset, we apply a difficulty filtering step using the Qwen3-4b-Instruct-2507~\citep{yang2025qwen3}. Specifically, for each question, we sample the model's responses 8 times, setting the temperature to 0.7 and the maximum token length to 6000. The statistics of the final filtered dataset are detailed in Appendix~\ref{app:final_data_statistics}.

For the exact instructions, please refer to the detailed prompt templates provided in Figure~\ref{fig:cyber_stage1_prompt}, ~\ref{fig:medical_stage1_prompt}, ~\ref{fig:cyber_stage3_prompt}, ~\ref{fig:medical_stage3_prompt}.

\subsection{Details of Cross-lingual Task Variant Synthesis}
\label{app:cs-detail}

\subsubsection{Details of Translation}
We synthesize the \textsc{Tgt-only} data by fully translating the \textsc{En-only} dataset. For the translation process, we utilize the Gemma-4-26B-A4B-it~\citep{gemma4_modelcard_2026} model configured with \texttt{thinking=off}. We translate the passages and multiple-choice options separately. This separate processing ensures the strict preservation of the \texttt{[MASK]} tokens and addresses the empirical tendency of large language models to under-translate or truncate long contexts. To mitigate this generation issue, passages are segmented and translated on a sentence-by-sentence basis, and the options are similarly split and translated individually. Furthermore, we provide few-shot exemplars within the prompt to enhance overall translation quality. Domain-specific terminology that must remain untranslated is managed separately via literal passthrough rules. The exact translation prompts are provided in Figure~\ref{fig:target_translation_prompt} and Figure~\ref{fig:options_translation_prompt}.

\subsubsection{Details of Code-switching Process}
We construct the option-side code-switching data by cross-matching the passages and multiple-choice options from the \textsc{Tgt-only} and \textsc{En-only} datasets. For the passage-level code-switched data, we utilize the Gemma-4-26B-A4B-it~\citep{gemma4_modelcard_2026} model configured with \texttt{thinking=off}. The code-switching pipeline operates by prompting the model to replace selected segments of a source passage with their target language counterparts. The model is strictly constrained to ensure contextual fluency, uniform span dispersion, and the exact preservation of \texttt{[MASK]} markers. The full prompt template is described at Figure~\ref{fig:code_switching_prompt}.

Following the generation, we enforce a post-hoc filtering step: if the cumulative character length of the switched spans constitutes less than 15\% of the total passage length, the generation is retried up to three times, with the decoding temperature increased by 0.3 for each sequential attempt.  

\input{latex/tables/final_data}

\subsection{Final data statistics}
\label{app:final_data_statistics}

Table~\ref{tab:final-data} summarizes the synthesized data pool by domain. We
refer to this as a synthesized pool rather than the final training set, because
our training runs use controlled subsamples and mix them with auxilary 
data as described in \S\ref{sec:experiments}.

\subsection{Qualitative Examples}
\label{app:qualitative_examples}
In this section, we present qualitative examples from our dataset. The examples for the medical domain are detailed in Figure~\ref{fig:qual-medical-1}, ~\ref{fig:qual-medical-2}, and those for the cybersecurity domain are provided in Figure~\ref{fig:qual-cybersecurity-1}, and ~\ref{fig:qual-cybersecurity-2}.

\section{Experimental Setup}
\label{app:setup}

\input{latex/tables/effective_samples}

\subsection{Additional Training Details}
\label{app:appendix-training-details}

\paragraph{Model-specific effective sample filtering.}
To isolate effective training instances without compromising response lengths, we apply a model-specific filtering procedure to the synthesized FIM-MCQA data pool for each domain--target-language pair. Specifically, for all training variants, we evaluate the respective models on the \textsc{En-only} data by sampling responses eight times per instance. We discard instances where the model answers correctly across all eight trials, as well as those with response lengths falling into the bottom 25th percentile. After applying this filtering process, we retain only the intersection of the surviving instances across all models to construct the final base dataset. The filtered data counts per model and domain are detailed in Table~\ref{tab:effective_samples}. 

\paragraph{Training data mixture.}
Each single-view variant utilizes the entirety of this filtered data for its corresponding language configuration. For multi-view variants, we sample uniformly across the constituent views so that the total contribution of each view is balanced. Note that the same FIM-MCQA instance preserves its answer index across its monolingual, bilingual, and code-switched realizations, enabling all variants to share the identical exact-match verifier.

\paragraph{Auxiliary General-Reasoning dataset.}
The auxiliary RLVR pool is shared across all RL-trained variants and consists
of $25\mathrm{K}$ scientific reasoning problems~\citep{akter2026nemotron},
$15\mathrm{K}$ math problems~\citep{deepscaler2025}, and $10\mathrm{K}$
instruction-following prompts~\citep{bercovich2025llamanemotronefficientreasoningmodels}.
For domain-RLVR variants, each batch samples prompts from the synthesized
domain FIM-MCQA pool with probability $0.6$ and from the auxiliary pool with
probability $0.4$. The \textsc{RLVR-Base} baseline is trained only on the auxiliary
pool. We use the same auxiliary pool and sampling rule across all domain
variants so that differences among \textsc{En-only}, \textsc{Tgt-only},
\textsc{Bi-Mono}, and \textsc{Ours} reflect the language realization of the
domain examples rather than changes in generic RLVR data.

\paragraph{DAPO hyperparameters.}
We post-train all RL variants with DAPO~\citep{yu2025dapo}. For each prompt,
we sample $N=16$ rollouts and apply accuracy-based group filtering, keeping a
group only when it contains at least one correct and one incorrect answer. We
disable the KL penalty and the overlong-buffer component. Rollouts are sampled
with temperature $0.6$ and top-$p=0.95$, using a maximum prompt length of
$4{,}200$ tokens and a maximum response length of $8{,}192$ tokens. We train
for two epoch with a batch size of $256$ prompts and a PPO mini-batch size of
$64$. Each variant requires approximately $2{,}500$ H100 GPU hours.

\paragraph{Validation and Evaluation.}
For cybersecurity, we reserve 10\% of CyberMetric for validation. For medicine,
we reserve an exclusive 10\% split of translated MedQA for validation. These validation
sets are used only for hyperparameter selection and are excluded from final
evaluation. The primary metric is pass@1 with $n=1$, temperature $T=0.6$, and top-$p=0.95$.

\subsection{Details of Evaluation Dataset Translation}
\label{app:eval_data_translation_details}
We utilize the GPT-5.4-mini~\citep{openai_gpt54_mini_2026}\footnote{Specifically, we used the \texttt{gpt-5.4-mini-2026-03-17} version.} model to perform the translation.
The prompt template for translating the evaluation data is provided in Figure~\ref{fig:eval_translation_prompt}. 

\input{latex/tables/translation_quality_comparison}
\input{latex/tables/medical_full}

\input{latex/tables/cyber_full}

\subsection{Translation Quality Assessment of the Evaluation Dataset}
\label{app:evaluation_dataset_translation_quality_evlaution}

To ensure the reliability of our cross-lingual evaluation, we assess the translation quality of the three cybersecurity benchmarks: SecEval \citep{li2023seceval}, CyberMetric \citep{tihanyi2024cybermetric}, and CTI-MCQ \citep{alam2024ctibench}. The evaluation process incorporates an LLM-as-a-judge approach, qualitative human assessment, and quantitative automated metrics.

To evaluate the translations comprehensively, we employ GPT-5.5 as an LLM judge to score the translated texts on a scale of 1 to 5 across three criteria: Accuracy, Fluency, and Terminology. Drawing inspiration from \citet{feng2025m}, we specifically adjusted our prompt to elicit explicit numerical scores (on a scale of 1 to 5) in a strict JSON format. This design choice enables a direct quantitative comparison to calculate the agreement between the LLM and human judgments. 
To validate the reliability of this LLM-based evaluation, we measure its alignment with human judgments.
We randomly subsampled 20 instances from each benchmark, and three independent evaluators with Computer Science backgrounds assessed them. To ensure a fair and consistent comparison between the machine and human assessments, both GPT-5.5~\citep{openai_gpt55_systemcard_2026} and the human experts were provided with the exact same instructions. The prompt utilized for this evaluation is illustrated in Figure \ref{fig:translate_quality_eval_prompt}. It asks the evaluator to output a strict JSON format scoring the translation on Accuracy, Fluency, and Terminology.

As shown in Table \ref{tab:translation_quality_comparison}, GPT-5.5 yielded high scores across all benchmarks. When compared to the human evaluation, we observed a strong positive correlation, with Pearson $r$ values ranging from 0.79 to 0.81 across the datasets. This significant alignment demonstrates that our LLM-based evaluation pipeline reliably mirrors human judgment, effectively verifying the translation quality.

In addition to the prompt-based evaluation, we perform a comprehensive automated evaluation using the COMET score \citep{rei2020comet} to measure the semantic alignment between the original English texts and their translated counterparts. Specifically, we utilize the \texttt{Unbabel/wmt22-cometkiwi-da} model and compute the scores for the full evaluation sets across both target languages, Korean and Swahili, to ensure broad quantitative validation.

The detailed results of the LLM evaluation, human assessment, and the COMET automated metric are summarized in Table \ref{tab:translation_quality_comparison}.

\section{Additional Evaluation Results}
\label{app:results}

\subsection{Full Evaluation Results}
\label{app:full_evaluation_results}
In this section, we report the full evaluation results, including the no-RLVR settings. The results for medical and cybersecurity domains are presented in Tables~\ref{tab:transposed-domain-ablation-v2} and \ref{tab:main-domain-ko-sw}.

\begin{table}[!t]
\centering
\footnotesize
\setlength{\tabcolsep}{3pt}
\newcommand{\sd}[2]{$#1_{\pm #2}$}
\newcommand{\sdb}[2]{$\mathbf{#1}_{\pm #2}$}
\begin{tabular}{@{}lcccc@{}}
\toprule
\textbf{Variant} & SecEval & CyberMetric & CTI-MCQ & \emph{Avg} \\
\midrule
\textsc{RLVR-Base} & \sd{43.9}{1.0} & \sd{65.5}{0.7} & \sd{40.5}{0.6} & \sd{50.0}{0.5} \\
\textsc{+En}       & \sd{51.5}{0.6} & \sd{70.6}{0.6} & \sd{43.8}{0.7} & \sd{55.3}{0.5} \\
\textsc{+Tgt}      & \sd{50.8}{0.5} & \sdb{80.0}{0.4} & \sd{45.2}{0.6} & \sd{58.7}{0.4} \\
\textsc{+BiMono}   & \sd{51.4}{0.4} & \sd{73.3}{0.4} & \sd{43.7}{0.6} & \sd{56.1}{0.5} \\
\textsc{+Ours}     & \sdb{57.5}{0.1} & \sdb{80.0}{0.4} & \sdb{46.9}{0.5} & \sdb{61.5}{0.2} \\
\bottomrule
\end{tabular}
\caption{Multi-seed results for SmolLM3-3B on Korean cybersecurity
benchmarks, evaluated in the target language. Values report mean $\pm$
standard deviation over three random seeds. Best results are bolded.}
\label{tab:multiseed-cyber}
\end{table}

\subsection{Multi-Seed Robustness}
\label{app:multiseed}

We extend the Korean cybersecurity comparison with SmolLM3-3B to three random seeds. Table~\ref{tab:multiseed-cyber} reports the mean and standard deviation for the five RL-trained variants. The same performance pattern is observed across the three seeds.
\textsc{+Ours} achieves the highest target-language average, performs best on
SecEval and CTI-MCQ, and matches \textsc{+Tgt} on CyberMetric.

\input{latex/tables/component_ablation}

\subsection{Detailed Component Ablation Results}
\label{app:ablation}
In this section, we present the component ablation study results across individual benchmarks. The complete results are detailed in Table~\ref{tab:component-ablation-main}.

\section{Analysis Details}
\label{app:analysis}

\input{latex/tables/consistency_table_perbench}

\subsection{Details of Cross-lingual Consistency Comparison}
\label{app:xling_consistency}
In this section, we present a detailed breakdown of the results from Table~\ref{tab:xling_consistency} across individual benchmarks. These results are provided in Table~\ref{tab:xling_consistency_perbench}.

\subsection{Details of In-domain Translation Experiment}
\label{app:gemba-da}
The performance ratio of variant $v$ on benchmark $b$ is
$\rho^{v,b}=\acc^{v,b}_{\tgt}/\acc^{v,b}_{\en}$, where $\acc^{v,b}_{\ell}$
is the variant's accuracy on the $N_b$ items of $b$ rendered in language
$\ell\in\{\en,\tgt\}$, evaluated on the same stem-key paired item set.
The translation prompts used by each model are provided in Figure~\ref{fig:model_gemba_translation_prompt}.
Figure~\ref{fig:gemba_prompt} shows the exact prompt template used for the LLM-based direct assessment (GEMBA-DA) with GPT-4.1~\citep{openai_gpt41_2025} to evaluate translation quality from the target language to English.

\input{latex/tables/full_pairwise_judge}

\subsection{Details of Representation Alignment Analysis}
\label{app:repr-details}

We use a fixed subset of $N=500$ semantically matched source--target question
pairs from CyberMetric. The source side is the original English question, and
the target side is the Korean translation used in our evaluation. For each
item, we render the same multiple-choice prompt in both languages and append
the same opening reasoning marker, \texttt{\textless think\textgreater}.

We run a single forward pass over the prompt plus the opening reasoning marker
and extract the hidden state at the marker position from every layer. If the
marker is split into multiple tokenizer tokens, we use the hidden state of the
final token of the marker. This position is immediately before the first
reasoning token would be predicted. Since the marker is identical across the
source and target prompts, differences in its contextualized hidden state
mainly reflect how the preceding question is represented before reasoning
begins, rather than variation from generated chain-of-thought continuations.

For each item $i$ and layer $\ell$, let $s_i^\ell$ and $t_i^\ell$ denote the
hidden states extracted from the source and target prompts. Stacking them over
all paired items gives layer-wise matrices
$S^\ell, T^\ell \in \mathbb{R}^{N \times d}$. We mean-center each language
separately over the item dimension:
\[
S_c^\ell = S^\ell - \bar{S}^\ell,
\qquad
T_c^\ell = T^\ell - \bar{T}^\ell.
\]

We compute two alignment metrics. First, Mean-Centered Cosine measures paired
item-level similarity after removing the per-language mean, which reduces the
effect of anisotropy in contextual LM representations
\citep{ethayarajh2019contextual,timkey2021rogue}:
\[
\mathrm{MCC}^{\ell}
=
\frac{1}{N}
\sum_{i=1}^{N}
\frac{
\langle S_{c,i}^{\ell}, T_{c,i}^{\ell} \rangle
}{
\lVert S_{c,i}^{\ell} \rVert
\lVert T_{c,i}^{\ell} \rVert
}.
\]

Second, ANC measures feature-level activation correspondence by averaging the
Pearson correlation of each hidden dimension over the aligned items
\citep{del-fishel-2022-cross}:
\[
\mathrm{ANC}^{\ell}
=
\frac{1}{d}
\sum_{j=1}^{d}
\mathrm{corr}
\left(
S_{c,:,j}^{\ell},
T_{c,:,j}^{\ell}
\right).
\]

We report the full layer-wise curves and summarize the results over a fixed
middle-layer band, layers 12--24 of the 36-layer model. The main text focuses
on this band because intermediate representations are commonly used to study
cross-lingual alignment, and because this range captures the largest separation
between the code-switched model and the non-code-switched baselines in our
analysis.

\subsection{Details of Generated Reasoning Trace Alignment}\label{app:appendix_judge}

For each benchmark item, we construct an English--target-language pair and collect outputs from each evaluated model under both language conditions. We compare \textsc{Ours} against \textsc{En-only} and \textsc{Tgt-only} using GPT-5.5 as a blind pairwise judge with reasoning effort set to \texttt{medium}. The judge receives anonymized outputs from two systems and selects the system whose English and target-language outputs are more semantically consistent in high-level logical progression, problem decomposition, intermediate conclusions, and solution strategy. The judge is instructed to ignore surface-level differences such as wording, formatting, verbosity, fluency, and language-specific phrasing.

To reduce position bias, we evaluate each comparison twice with the A/B order reversed. We aggregate the two judgments into a single outcome per example. If both orders prefer the same underlying system, we count it as a win for that system. We count an example as a tie only when both judgments explicitly return a tie. Unless otherwise specified, $n$ denotes the number of examples before A/B order swapping.

For the medical domain, we evaluate Qwen2.5-3B-Instruct and Llama-3.2-3B-Instruct on four benchmarks with 100 randomly sampled examples per benchmark, resulting in $n=400$ examples per model family. For cybersecurity, we evaluate SmolLM3-3B on three benchmarks with 100 examples per benchmark, resulting in $n=300$. The judge prompt is shown in Figure~\ref{fig:judge_prompt}, and the per-benchmark results are reported in Table~\ref{tab:judge_full_appendix}.

\section{Prompts}
\label{app:prompts}
In this section, we present the full prompt templates used in our experiments. To facilitate navigation, we group these prompts according to their specific tasks: extraction synthesis, translation, code-switching, and evaluation.

\subsection{Extraction Synthesis Prompts}
\label{app:prompts-extraction}
The prompts for extracting educational passages and generating distractors across domains are provided in Figures~\ref{fig:cyber_stage1_prompt}, \ref{fig:medical_stage1_prompt}, \ref{fig:cyber_stage3_prompt}, and \ref{fig:medical_stage3_prompt}.

\subsection{Translation Prompts}
\label{app:prompts-translation}
The prompts utilized for context-aware passage translation, option translation, evaluation benchmark translation, and direct translation are shown in Figures~\ref{fig:target_translation_prompt}, \ref{fig:options_translation_prompt}, \ref{fig:eval_translation_prompt}, and \ref{fig:model_gemba_translation_prompt}.

\subsection{Code-Switching Prompt}
\label{app:prompts-cs}
The prompt template used for span-level code-switching generation is detailed in Figure~\ref{fig:code_switching_prompt}.

\subsection{Evaluation and Judging Prompts}
\label{app:prompts-eval}
The prompts used for evaluation tasks, including GEMBA-DA, reasoning trace alignment, and translation quality assessment, are presented in Figures~\ref{fig:gemba_prompt}, \ref{fig:judge_prompt}, and \ref{fig:translate_quality_eval_prompt}.

\section{Usage of AI assistants}
\label{app:ai-usage}
During the preparation of this manuscript, AI-based writing tools were used solely to improve the clarity, grammar, and readability of the English text. They were not involved in formulating the research questions, developing the methodology, conducting the experiments, or interpreting the results. All technical content, analyses, and conclusions were independently developed by the authors.

\newpage
\input{latex/figures/qual_med_eg_1}
\newpage
\input{latex/figures/qual_med_eg_2}
\newpage
\input{latex/figures/qual_cyber_eg_1}
\newpage
\input{latex/figures/qual_cyber_eg_2}

\begin{figure*}[htbp]
\begin{tcolorbox}[colback=gray!5, colframe=gray!50, title=Extraction Synthesis Stage 1 Prompt (Cybersecurity)]
\begin{lstlisting}[breaklines=true, basicstyle=\ttfamily\small]
Given a source document, perform TWO tasks:

1. **Extract an educational passage**: Extract a coherent, self-contained passage (around 200-300 words) from the document that is educationally valuable for learning Cybersecurity. The passage should contain reasoning - causal explanations, attack mechanism analysis, vulnerability consequences, or defensive logic. Remove noise such as ads, navigation menus, author bylines, dates, and promotional content. The passage must read as a clean, standalone educational text.

2. **Identify the mask span**: Within the extracted passage, identify a consecutive multi-sentence span (around 50-100 words) that contains the most important reasoning steps. This span will be masked and serve as the ground-truth answer.

VERBATIM COPYING RULES - violating these causes a task failure:
1. Both "passage" and "removed_steps" must be copied CHARACTER-FOR-CHARACTER from the original document.
2. Do NOT paraphrase, simplify, reorder, or summarize.
3. Do NOT add or remove spaces, line breaks, or punctuation.
4. "removed_steps" must be a substring of "passage" - every character must appear in the passage exactly as written.

Return ONLY the following JSON within <answer></answer> tags:

<answer>
{
  "passage": "...",
  "removed_steps": "..."
}
</answer>

If there is no suitable educational passage about Cybersecurity, or if the document is not written in English, return an empty string within <answer></answer>.

Document:
SOLUTION_PLACEHOLDER
\end{lstlisting}
\end{tcolorbox}
\caption{The prompt template used in Stage 1 to simultaneously extract educational passages and identify mask spans for the Cybersecurity domain.}
\label{fig:cyber_stage1_prompt}
\end{figure*}
 
\begin{figure*}[htbp]
\begin{tcolorbox}[colback=gray!5, colframe=gray!50, title=Extraction Synthesis Stage 1 Prompt (Medical)]
\begin{lstlisting}[breaklines=true, basicstyle=\ttfamily\small]
Given a source document, perform TWO tasks:

1. **Extract an educational passage**: Extract a coherent, self-contained passage (around 200-300 words) from the document that is educationally valuable for learning Medicine or Biomedical Science. The passage should contain reasoning - clinical reasoning, pathophysiology, mechanism of disease, diagnostic logic, treatment rationale, or pharmacological mechanisms. Remove noise such as ads, navigation menus, author bylines, dates, and promotional content. The passage must read as a clean, standalone educational text.

2. **Identify the mask span**: Within the extracted passage, identify a consecutive multi-sentence span (around 50-100 words) that contains the most important reasoning steps. This span will be masked and serve as the ground-truth answer.

VERBATIM COPYING RULES - violating these causes a task failure:
1. Both "passage" and "removed_steps" must be copied CHARACTER-FOR-CHARACTER from the original document.
2. Do NOT paraphrase, simplify, reorder, or summarize.
3. Do NOT add or remove spaces, line breaks, or punctuation.
4. "removed_steps" must be a substring of "passage" - every character must appear in the passage exactly as written.

Return ONLY the following JSON within <answer></answer> tags:

<answer>
{
  "passage": "...",
  "removed_steps": "..."
}
</answer>

If there is no suitable educational passage about Medicine or Biomedical Science, or if the document is not written in English, return an empty string within <answer></answer>.

Document:
SOLUTION_PLACEHOLDER
\end{lstlisting}
\end{tcolorbox}
\caption{The prompt template used in Stage 1 to simultaneously extract educational passages and identify mask spans for the Medical domain.}
\label{fig:medical_stage1_prompt}
\end{figure*}

\begin{figure*}[htbp]
\begin{tcolorbox}[colback=gray!5, colframe=gray!50, title=Extraction Synthesis Stage 3 Prompt (Cybersecurity)]
\begin{lstlisting}[breaklines=true, basicstyle=\ttfamily\small]
You are given a cybersecurity document, its masked version showing where the [MASK] was removed, and the removed span. Generate at least ten diverse distractors that are plausible, similar in style to the removed content, but incorrect.

CRITICAL: Each distractor MUST match the word count of the removed span. Count the words in the removed span and ensure every distractor has a similar word count.

Return your answer in the following JSON format within <answer></answer>:

<answer>
{
  "distractors": [...]
}
</answer>

Document:
SOLUTION_PLACEHOLDER

Document with [MASK]:
MASKED_SOLUTION_PLACEHOLDER

Removed span (correct answer):
REMOVED_STEPS_PLACEHOLDER
\end{lstlisting}
\end{tcolorbox}
\caption{The prompt template used in Stage 3 to generate plausible, length-matched distractors for the masked spans in the Cybersecurity domain.}
\label{fig:cyber_stage3_prompt}
\end{figure*}

\begin{figure*}[htbp]
\begin{tcolorbox}[colback=gray!5, colframe=gray!50, title=Extraction Synthesis Stage 3 Prompt (Medical)]
\begin{lstlisting}[breaklines=true, basicstyle=\ttfamily\small]
You are given a medical document, its masked version showing where the [MASK] was removed, and the removed span. Generate at least ten diverse distractors that are plausible, similar in style to the removed content, but incorrect.

CRITICAL: Each distractor MUST match the word count of the removed span. Count the words in the removed span and ensure every distractor has a similar word count.

Return your answer in the following JSON format within <answer></answer>:

<answer>
{
  "distractors": [...]
}
</answer>

Document:
SOLUTION_PLACEHOLDER

Document with [MASK]:
MASKED_SOLUTION_PLACEHOLDER

Removed span (correct answer):
REMOVED_STEPS_PLACEHOLDER
\end{lstlisting}
\end{tcolorbox}
\caption{The prompt template used in Stage 3 to generate plausible, length-matched distractors for the masked spans in the Medical domain.}
\label{fig:medical_stage3_prompt}
\end{figure*}

\begin{figure*}[t]
\begin{tcolorbox}[colback=gray!5, colframe=gray!50, title=Context-Aware Translation Prompt]
\begin{Verbatim}[fontsize=\small, breaklines=true, breakanywhere=true]

Using the {domain} document below as context, translate the given text into {language} ({native}).
Translation rules:
- Translate the **entire text** into {language}. Even if the text is long, translate it all the way through.
- Do not leave any English sentence and do not break off the translation midway.
- Keep ONLY {keep_in_english} in English. All other English text must be **fully** translated.
- {literal_passthroughs} remain unchanged, but the surrounding sentences must be translated.

- Never modify the markers [MASK].
- Do NOT switch language mid-word (no English suffix glued to a {language} stem, no {language} character inserted inside an English word). A technical term must appear in exactly one of these forms: (a) all English, (b) all {language}, or (c) {language} followed by the English term in parentheses.

- Use consistent {language} technical terminology across the whole text.
- Return only the translated text, with no explanation or extra quotation marks.

Translation example:
Input:  "{ex_in_single}"
Output: "{ex_out_single}"

Document (translation context):
PASSAGE_PLACEHOLDER

Text to translate:
TEXT_PLACEHOLDER
\end{Verbatim}
\end{tcolorbox}
\caption{The prompt template used for context-aware translation into the target language. The prompt enforces strict constraints to preserve code snippets, system identifiers, and \texttt{[MASK]} markers while preventing unnatural mid-word language mixing.}
\label{fig:target_translation_prompt}
\end{figure*}

\begin{figure*}[t]
\begin{tcolorbox}[colback=gray!5, colframe=gray!50, title=Context-Aware Options Translation Prompt]
\begin{Verbatim}[fontsize=\small, breaklines=true, breakanywhere=true]
Using the {domain} document below as context, translate the given options into {language} ({native}).

Translation rules:
- Translate the **entire content of all options** into {language}. Do not translate only part of an option, and do not leave any sentence in English.
- Even if an option is long, translate it through to the end without stopping.
- {literal_passthroughs} remain in their original form, but the surrounding sentences must be in {language}.
- Never modify the markers [MASK].
- Do NOT switch language mid-word (no English suffix on a {language} stem, and no {language} character inside an English word). A term must be (a) all English, (b) all {language}, or (c) {language} followed by English in parentheses.
- Use consistent {language} technical terminology across all options.
- Return only the translated option text, with no commentary or metadata.

Translation example:
Input:  "{ex_in_option}"
Output: "{ex_out_option}"

Return the following JSON format inside <answer></answer> tags:

<answer>
{
  "translated": ["<option_1>", "<option_2>", ...]
}
</answer>

Document (translation context):
PASSAGE_PLACEHOLDER

Options to translate:
OPTIONS_PLACEHOLDER
\end{Verbatim}
\end{tcolorbox}
\caption{The prompt template used to translate the multiple-choice options into the target language. Similar to the passage translation, it leverages the source document as context to ensure terminology consistency and requires a strict JSON array output.}
\label{fig:options_translation_prompt}
\end{figure*}

\begin{figure*}[t]
\begin{tcolorbox}[colback=gray!5, colframe=gray!50, title=Evaluation Benchmark Translation Prompt]
\begin{Verbatim}[fontsize=\small, breaklines=true, breakanywhere=true]
You are a professional {language} translator specializing in {domain}.
Translate the following {domain} multiple-choice question and its options into natural {language} ({native}).

Rules:
- Translate naturally and fluently into {language} - readers should feel the text was originally written in {language}.
- Do NOT translate or modify any code blocks, command-line examples, or inline code snippets.
- Keep option labels (A), B), C), D) etc.) exactly as-is at the start of each option.
- Output ONLY the translated question and options, nothing else.

Text to translate:
{text}
\end{Verbatim}
\end{tcolorbox}
\caption{The prompt template used to translate evaluation benchmark multiple-choice questions and options into the target language.}
\label{fig:eval_translation_prompt}
\end{figure*}

\begin{figure*}[t]
\begin{tcolorbox}[colback=gray!5, colframe=gray!50, title=Direct Translation Prompt]
\begin{Verbatim}[fontsize=\small, breaklines=true, breakanywhere=true]
Translate the following {src_lang_full} text to English. 
Output only the English translation, no explanation, no preamble.

{src_lang_full} text:
{source_text}
\end{Verbatim}
\end{tcolorbox}
\caption{The prompt template used for translating source language text into English.}
\label{fig:model_gemba_translation_prompt}
\end{figure*}

\begin{figure*}[t]
\begin{tcolorbox}[colback=gray!5, colframe=gray!50, title=Span-Level Code-Switching Generation Prompt]
\begin{Verbatim}[fontsize=\small, breaklines=true, breakanywhere=true]
You will code-switch a {source} ({source_native}) {domain} text into {target} ({target_native}) by replacing selected spans with their {target} translations. Choose spans so that the resulting text READS NATURALLY - the output should feel like a fluent bilingual {domain} writer's prose, not a mechanical word-by-word substitution.

Rules (STRICT):
1. Pick spans whose {target} translation READS NATURALLY in context. Translate content-carrying spans - nouns, verbs, adjectives, short noun/verb phrases - whenever their {target} rendering is smooth. Spans may be single words or short 2-4 token phrases - phrasal translations often read more naturally than single-word swaps, so prefer them when they flow better.
2. Domain literals stay in their original form (do NOT translate or transliterate): {literal_passthroughs}, code blocks, equations.
3. Spread selections EVENLY across the whole passage - the beginning, the middle, and the end must each contribute some switched spans. Clustering in one region is a failure.
4. If the same word appears several times, you do NOT have to translate every occurrence. Translating one or two occurrences and leaving the others in {source} is actually preferred - it looks more natural and avoids the "always the same word" bias.
5. Translate each selected span into natural {target} and substitute it in place of the {source} original. Keep the surrounding {source} text exactly as it was (same whitespace, same casing, same punctuation, same paragraph breaks).
6. NEVER modify, translate, remove, or reorder the markers [MASK]. They must appear the same number of times and in the same order in your output.
7. Do not add, remove, or reorder sentences. Do not paraphrase untouched text.

{example_block}
OUTPUT FORMAT - respond with ONLY the following, nothing else:
<answer>
{
  "code_switched": "<the full passage after code-switching>",
  "switched_tokens": ["<{source} span 1>", "<{source} span 2>", ...]
}
</answer>

Emit exactly ONE opening `{` and ONE matching closing `}`. Do NOT double them.
Escape newlines inside the JSON string as \n and backslashes as \\.

PASSAGE:
<<<
PASSAGE_PLACEHOLDER
>>>
\end{Verbatim}
\end{tcolorbox}
\caption{The prompt template used for generating span-level code-switched passages.}
\label{fig:code_switching_prompt}
\end{figure*}

\begin{figure*}[htbp]
\begin{tcolorbox}[colback=gray!5, colframe=gray!50, title=GEMBA-DA Evaluation Prompt]
\begin{lstlisting}[breaklines=true, basicstyle=\ttfamily\small]
Score the following translation from {src_full} to {tgt_full} on a 
continuous scale from 0 to 100, where a score of zero means "no meaning 
preserved" and score of one hundred means "perfect meaning and grammar".

{src_full} source: "{src}"
{tgt_full} translation: "{tgt}"

Score (0-100):
\end{lstlisting}
\end{tcolorbox}
\caption{The prompt template for GEMBA-DA used to measure translation quality. The variables \texttt{\{src\_full\}} and \texttt{\{tgt\_full\}} represent the full names of the source and target languages, while \texttt{\{src\}} and \texttt{\{tgt\}} are substituted with the original text and the model's translation, respectively.}
\label{fig:gemba_prompt}
\end{figure*}

\begin{figure*}[htbp]
\begin{tcolorbox}[colback=gray!5, colframe=gray!50, title=Reasoning Trace Alignment Judge Prompt]
\begin{Verbatim}[fontsize=\small, breaklines=true, breakanywhere=true]
You will see TWO models' reasoning behavior on the same problem.
Each model produced one English trace (for the English version 
of the problem) and one {target_name} trace (for the 
{target_name} version of the same problem).

Your task: judge which MODEL shows MORE CONSISTENT REASONING 
between its English and {target_name} responses.

If the two models show a similar degree of reasoning consistency, choose TIE. Do not force a choice between the two models when there is no meaningful difference in consistency.

WHAT "CONSISTENT" MEANS:
- The model decomposes the problem the same way in both languages
- Uses similar key facts, concepts, formulas across both traces
- Follows a similar high-level strategy in both languages
- Arrives at similar intermediate insights

WHAT "CONSISTENT" DOES NOT MEAN:
- Both traces being correct (a model can be consistently wrong, 
  and that still counts as consistent)
- The {target_name} trace using English words (this is a surface 
  feature, not consistency)
- Either trace being more detailed or longer than the other
- Either model being individually "smarter" or producing 
  higher-quality reasoning

PROBLEM (English):
{Q_en}

PROBLEM ({target_name}):
{Q_tgt}
\end{Verbatim}
\end{tcolorbox}
\caption{The prompt template used for the reasoning trace alignment experiment.}
\label{fig:judge_prompt}
\end{figure*}

\begin{figure*}[h]
\begin{tcolorbox}[colback=gray!5, colframe=gray!50, title=Evaluation Data Translation Quality Assessment Prompt]
\begin{lstlisting}[breaklines=true, basicstyle=\ttfamily\small]
Evaluate the translation on three aspects (1=worst, 5=best):
1. Accuracy: Does the translation preserve the source meaning?
2. Fluency: Is the translation grammatically correct and natural?
3. Terminology: Are domain-specific terms translated correctly?
Domain: {domain}
Source: {src}
Translation: {tgt}
Output JSON only:
{"accuracy": <1-5>, "fluency": <1-5>, "terminology": <1-5>}
\end{lstlisting}
\end{tcolorbox}
\caption{The prompt template used for both the LLM-as-a-judge (GPT-5.5) and the human evaluators to assess translation quality.}
\label{fig:translate_quality_eval_prompt}
\end{figure*}

\end{document}

%% file: latex/tables/medical_main.tex
\begin{table*}[t]
\centering
\setlength{\tabcolsep}{4.5pt}
\definecolor{pointgreen}{HTML}{03C75A}
\resizebox{\textwidth}{!}{
\begin{tabular}{@{} l ccccc | ccccc @{}}
\toprule
& \multicolumn{5}{c|}{\textbf{Qwen2.5-3B-Instruct}} & \multicolumn{5}{c}{\textbf{Llama-3.2-3B-Instruct}} \\
\cmidrule(lr){2-6} \cmidrule(lr){7-11}
\textbf{Bench.} & RLVR-Base & +En & +Tgt & +BiMono & +Ours & RLVR-Base & +En & +Tgt & +BiMono & +Ours \\
\midrule
KorMedMQ-D & 35.06 & 35.34 & 35.29 & 35.34 & \cellcolor{pointgreen!15}\textbf{36.67} & 21.10 & 27.93 & 27.47 & 28.79 & \cellcolor{pointgreen!15}\textbf{29.31} \\
MedMCQA* & 34.52 & 33.96 & 34.03 & 33.72 & \cellcolor{pointgreen!15}\textbf{34.86} & 24.26 & 27.12 & 27.36 & \textbf{29.65} & \cellcolor{pointgreen!15}29.22 \\
KMMLU-Pro-M & 35.45 & 35.76 & 36.21 & 36.06 & \cellcolor{pointgreen!15}\textbf{36.29} & 22.84 & 24.77 & 24.09 & 25.08 & \cellcolor{pointgreen!15}\textbf{26.36} \\
\cmidrule(lr){1-11}
\textit{Avg (Tgt)} & 35.01 & 35.02 & 35.18 & 35.04 & \cellcolor{pointgreen!25}\textbf{35.94} & 22.73 & 26.61 & 26.31 & 27.84 & \cellcolor{pointgreen!25}\textbf{28.30} \\
\color{gray}\textit{Avg (En)} & \color{gray} 51.42 & \color{gray} 52.19 & \color{gray} \textbf{52.97} & \color{gray} 52.96 & \color{gray} 52.13 & \color{gray} 51.65 & \color{gray} 56.64 & \color{gray} 56.89 & \color{gray} 57.21 & \color{gray} \textbf{57.26} \\
\bottomrule
\end{tabular}
}
\caption{
Performance on medical benchmarks across RLVR data configurations.
\textsc{RLVR-Base} denotes RLVR training using only the auxiliary data pool.
The remaining variants include the same auxiliary pool and additionally use English data (\textsc{+En}), target-language data (\textsc{+Tgt}), bilingual monolingual data (\textsc{+BiMono}), or our proposed code-switched data (\textsc{+Ours}).
The asterisk (*) indicates that the benchmark is translated into the respective language.
The gray \textit{Avg (En)} row reports average English performance (details in Appendix~\ref{app:full_evaluation_results}).
Bold indicates the best result within each model, and green shading highlights our proposed method (\textsc{+Ours}).
}
\label{tab:transposed-domain-ablation-v3}
\end{table*}

%% file: latex/tables/cyber_main.tex
\begin{table*}[t]
\centering
\setlength{\tabcolsep}{4.5pt}
\definecolor{pointgreen}{HTML}{03C75A}
\resizebox{\textwidth}{!}{
\begin{tabular}{@{} l ccccc | ccccc @{}}
\toprule
& \multicolumn{10}{c}{\textbf{SmolLM3-3B}} \\
\cmidrule(lr){2-11}
& \multicolumn{5}{c|}{\textbf{Korean}} & \multicolumn{5}{c}{\textbf{Swahili}} \\
\cmidrule(lr){2-6} \cmidrule(lr){7-11}
\textbf{Bench.} & RLVR-Base & +En & +Tgt & +BiMono & +Ours & RLVR-Base & +En & +Tgt & +BiMono & +Ours \\
\midrule
SecEval* & 44.3 & 50.1 & 49.8 & 51.1 & \cellcolor{pointgreen!15}\textbf{57.3} & 25.8 & 26.8 & \textbf{28.6} & 25.9 & \cellcolor{pointgreen!15}28.3 \\
CyberMetric* & 65.8 & 70.6 & \textbf{80.2} & 72.4 & \cellcolor{pointgreen!15}80.0 & 54.0 & 53.9 & 54.2 & 54.5 & \cellcolor{pointgreen!15}\textbf{55.0} \\
CTI-MCQ* & 40.6 & 43.2 & 45.5 & 43.8 & \cellcolor{pointgreen!15}\textbf{47.1} & 41.9 & 42.1 & 43.2 & 43.4 & \cellcolor{pointgreen!15}\textbf{45.0} \\
\cmidrule(lr){1-11}
\textit{Avg (Tgt)} & 50.23 & 54.63 & 58.50 & 55.77 & \cellcolor{pointgreen!25}\textbf{61.47} & 40.57 & 40.93 & 42.00 & 41.27 & \cellcolor{pointgreen!25}\textbf{42.77} \\
\midrule
& \multicolumn{5}{c|}{\textbf{Hindi}} & \multicolumn{5}{c}{\textbf{Japanese}} \\
\cmidrule(lr){2-6} \cmidrule(lr){7-11}
\textbf{Bench.} & RLVR-Base & +En & +Tgt & +BiMono & +Ours & RLVR-Base & +En & +Tgt & +BiMono & +Ours \\
\midrule
SecEval* & 39.2 & 54.6 & 54.9 & 53.1 & \cellcolor{pointgreen!15}\textbf{56.0} & 50.6 & 51.3 & 52.2 & 52.0 & \cellcolor{pointgreen!15}\textbf{53.4} \\
CyberMetric* & 66.7 & 71.2 & 73.5 & 74.5 & \cellcolor{pointgreen!15}\textbf{77.0} & 77.4 & 77.4 & 78.4 & 77.8 & \cellcolor{pointgreen!15}\textbf{79.8} \\
CTI-MCQ* & 48.0 & 50.9 & \textbf{51.4} & 50.7 & \cellcolor{pointgreen!15}\textbf{51.4} & 49.7 & 51.1 & 51.2 & 52.0 & \cellcolor{pointgreen!15}\textbf{52.3} \\
\cmidrule(lr){1-11}
\textit{Avg (Tgt)} & 51.30 & 58.90 & 59.93 & 59.43 & \cellcolor{pointgreen!25}\textbf{61.47} & 59.23 & 59.93 & 60.60 & 60.60 & \cellcolor{pointgreen!25}\textbf{61.83} \\
\bottomrule
\end{tabular}
}
\caption{
Comprehensive performance on cybersecurity benchmarks across RLVR data configurations for four target languages spanning different scripts and resource levels: Korean and Swahili (top), and Hindi and Japanese (bottom). All other notations and formatting follow Table~\ref{tab:transposed-domain-ablation-v3}.}
\label{tab:main-domain-ko-sw_v2}
\end{table*}

%% file: latex/tables/component_ablation_compact.tex
\begin{table}[t]
\centering
\small
\definecolor{pointgreen}{HTML}{03C75A}
\begin{tabular*}{\linewidth}{@{\extracolsep{\fill}} l cc}
\toprule
\textbf{Variant} & \textbf{Target Avg} & \textbf{English Avg} \\
\midrule
RLVR-Base        & 50.23 & 64.87 \\
\quad + En       & 54.63 & 67.00 \\
\quad + Tgt      & 58.50 & 67.83 \\
\quad + BiMono   & 55.77 & 66.80 \\
\quad + Opt      & 56.30 & 67.20 \\
\quad + Pass     & 59.37 & 67.07 \\
\quad + Ours     & \textbf{61.47} & \textbf{68.27} \\
\bottomrule
\end{tabular*}
\caption{Component ablation on Korean cybersecurity benchmarks (SmolLM3-3B). Values report the macro-average accuracy across benchmarks in the Target Language (Korean) and English.}
\label{tab:component-ablation-compact}
\end{table}

%% file: latex/tables/consistency_table_main.tex
\begin{table}[t]
\centering
\small
\begin{tabular}{lccc}
\toprule
(English, Korean) & ($\checkmark$,$\checkmark$) $\uparrow$ & ($\checkmark$,$\times$) $\downarrow$ & ($\times$,$\checkmark$) $\downarrow$ \\
\midrule
Base & 15.1 & 28.2 & 13.0 \\
En-only & 28.1 & 30.0 & 7.3 \\
Bi-Mono & 29.2 & 29.8 & 7.1 \\
\textbf{Ours} & \textbf{34.8} & \textbf{27.1} & \textbf{6.7} \\
\bottomrule
\end{tabular}
\caption{Cross-lingual consistency comparison on cybersecurity benchmarks. This table aggregates the outcomes of English and Korean question pairs across cybersecurity domain benchmarks. $\checkmark$ denotes a correct answer, and $\times$ denotes an incorrect answer. \textbf{Bold} indicates the most consistent results.}
\label{tab:xling_consistency}
\end{table}

%% file: latex/tables/pairwise_judge_main.tex
\begin{table}[t]
\centering
\small
\begin{tabular}{@{} l cc cc @{}}
\toprule
& \multicolumn{2}{c}{vs.\ \textsc{En-only}} & \multicolumn{2}{c}{vs.\ \textsc{Tgt-only}} \\
\cmidrule(lr){2-3}\cmidrule(l){4-5}
Win Rate(\%) & Ours & En-only & Ours & Tgt-only \\
\midrule
\multicolumn{5}{@{}l}{\textbf{Medical}} \\
\cmidrule(lr){1-5}
Qwen2.5   & \textbf{48.3} & 40.0 & \textbf{47.5} & 42.5 \\
Llama3.2  & \textbf{47.0} & 38.8 & \textbf{43.3} & 42.0 \\
\midrule
\multicolumn{5}{@{}l}{\textbf{Cyber}} \\
\cmidrule(lr){1-5}
SmolLM3   & \textbf{36.3} & 31.7 & \textbf{58.7} & 21.7 \\
\bottomrule
\end{tabular}
\caption{Pairwise judge win rates to measure cross-lingual reasoning consistency using GPT-5.5. The target language is Korean. The reported results are aggregated across benchmarks within each domain (evaluating 100 samples per benchmark). Our method demonstrates consistent improvement across all models and baselines.}
\label{tab:judge_main}
\end{table}

%% file: latex/tables/final_data.tex
\begin{table}[t]
\centering
\small
\begin{tabular}{@{}lrl@{}}
\toprule
\textbf{Domain} & \textbf{$n$} & \textbf{Source} \\
\midrule
Cybersecurity & 207{,}048 & Primus-Seed, Primus-FineWeb \\
Medical       & 158{,}714 & Biomed-Enriched \\
\midrule
\textbf{Total} & \textbf{365{,}762} &  \\
\bottomrule
\end{tabular}
\caption{
Synthesized FIM-MCQA data pool per domain. Note that for each domain-target language pair, an identical number of samples is constructed across all training variants, matching the exact counts reported in the table. The subsample used for
training is described in Appendix~\ref{app:appendix-training-details}.
}
\label{tab:final-data}
\end{table}

%% file: latex/tables/effective_samples.tex
\begin{table}[t]
\centering
\setlength{\tabcolsep}{6pt}
\resizebox{\linewidth}{!}{
\begin{tabular}{llrr}
\toprule
Domain & Model & Pool & Filtered \\
\midrule
Medical & Qwen2.5-3B-Instruct & 158,714 & \textbf{50,476} \\
Medical & Llama-3.2-3B-Instruct & 158,714 & \textbf{44,294} \\
\midrule
Cybersecurity & SmolLM3-3B & 207,048 & \textbf{43,705} \\
\bottomrule
\end{tabular}
}
\caption{Number of Effective Samples after the difficulty filtering stage.}
\label{tab:effective_samples}
\end{table}

%% file: latex/tables/translation_quality_comparison.tex
\begin{table}[h]
\centering
\resizebox{\linewidth}{!}{
\setlength{\tabcolsep}{4pt} %
\begin{tabular}{@{} l c c c c c @{}}
\toprule
\multirow{2}{*}{\textbf{Benchmark}} & \multicolumn{3}{c}{\textbf{Korean (Acc/Flu/Term)}} & \multicolumn{2}{c}{\textbf{COMET}} \\
\cmidrule(lr){2-4} \cmidrule(lr){5-6}
& \textbf{GPT-5.5} & \textbf{Human} & \textbf{$r$} & \textbf{KO} & \textbf{SW} \\
\midrule
SecEval     & 4.94/4.82/4.70 & 4.25/4.18/4.12 & 0.81 & 0.852 & 0.803 \\
CyberMetric & 4.92/4.75/4.67 & 4.21/4.14/4.08 & 0.83 & 0.864 & 0.810 \\
CTI-MCQ     & 4.88/4.74/4.66 & 4.16/4.10/4.02 & 0.79 & 0.851 & 0.808 \\
\bottomrule
\end{tabular}
}
\caption{Translation quality evaluation across cybersecurity benchmarks. GPT-5.5 and human scores represent assessments on a 20-instance subsample for Korean (evaluated on a 1-5 scale). $r$ denotes the Pearson correlation between LLM and human judgments. COMET scores are reported for the full evaluation sets in Korean and Swahili.}
\label{tab:translation_quality_comparison}
\end{table}

%% file: latex/tables/medical_full.tex
\begin{table*}[!t]
\centering
\small
\setlength{\tabcolsep}{4.5pt}
\definecolor{pointgreen}{HTML}{03C75A}

\resizebox{\textwidth}{!}{ 
\begin{tabular}{@{} l cccccc | cccccc @{}}
\toprule
& \multicolumn{6}{c|}{\textbf{Qwen2.5-3B-Instruct}} & \multicolumn{6}{c}{\textbf{Llama-3.2-3B-Instruct}} \\
\cmidrule(lr){2-7} \cmidrule(lr){8-13}
\textbf{Bench.} & Base & RLVR-Base & +En & +Tgt & +BiMono & +Ours & Base & RLVR-Base & +En & +Tgt & +BiMono & +Ours \\
\midrule

\multicolumn{13}{@{}l}{\textbf{Target Language}} \\
\midrule
KorMedMQ-D & 34.50 & 35.06 & 35.34 & 35.29 & 35.34 & \cellcolor{pointgreen!15}\textbf{36.67} & 19.02 & 21.10 & 27.93 & 27.47 & 28.79 & \cellcolor{pointgreen!15}\textbf{29.31} \\
MedMCQA* & 33.24 & 34.52 & 33.96 & 34.03 & 33.72 & \cellcolor{pointgreen!15}\textbf{34.86} & 19.85 & 24.26 & 27.12 & 27.36 & \textbf{29.65} & \cellcolor{pointgreen!15}29.22 \\
KMMLU-Pro-M & 35.20 & 35.45 & 35.76 & 36.21 & 36.06 & \cellcolor{pointgreen!15}\textbf{36.29} & 20.45 & 22.84 & 24.77 & 24.09 & 25.08 & \cellcolor{pointgreen!15}\textbf{26.36} \\
\cmidrule(lr){1-13}
\textit{Avg (Tgt)} & 34.31 & 35.01 & 35.02 & 35.18 & 35.04 & \cellcolor{pointgreen!25}\textbf{35.94} & 19.77 & 22.73 & 26.61 & 26.31 & 27.84 & \cellcolor{pointgreen!25}\textbf{28.30} \\
\midrule

\multicolumn{13}{@{}l}{\color{gray} \textbf{English}} \\
\midrule
\color{gray} KorMedMQ-D* & \color{gray} 57.53 & \color{gray} 57.74 & \color{gray} 58.05 & \color{gray} \textbf{61.03} & \color{gray} 60.06 & \color{gray} 58.74 & \color{gray} 48.85 & \color{gray} 50.01 & \color{gray} 56.72 & \color{gray} 57.30 & \color{gray} 57.53 & \color{gray} \textbf{57.93} \\
\color{gray} MedMCQA & \color{gray} 47.36 & \color{gray} \textbf{47.88} & \color{gray} 47.60 & \color{gray} 47.42 & \color{gray} 47.34 & \color{gray} 47.01 & \color{gray} 52.28 & \color{gray} 52.57 & \color{gray} 55.00 & \color{gray} 55.00 & \color{gray} \textbf{55.17} & \color{gray} 55.04 \\
\color{gray} MedQA & \color{gray} 50.29 & \color{gray} 48.65 & \color{gray} 50.92 & \color{gray} 50.47 & \color{gray} \textbf{51.47} & \color{gray} 50.65 & \color{gray} 50.63 & \color{gray} 52.38 & \color{gray} 58.19 & \color{gray} 58.37 & \color{gray} \textbf{58.92} & \color{gray} 58.82 \\
\cmidrule(lr){1-13}
\color{gray} \textit{Avg (En)} & \color{gray} 51.73 & \color{gray} 51.42 & \color{gray} 52.19 & \color{gray} \textbf{52.97} & \color{gray} 52.96 & \color{gray} 52.13 & \color{gray} 50.59 & \color{gray} 51.65 & \color{gray} 56.64 & \color{gray} 56.89 & \color{gray} 57.21 & \color{gray} \textbf{57.26} \\

\bottomrule
\end{tabular}
}
\caption{
Comprehensive performance on medical benchmarks across RLVR data configurations.
\textsc{Base} denotes the original instruction-tuned model without RLVR.
\textsc{RLVR-Base} denotes RLVR training using only the auxiliary data pool.
The remaining variants include the same auxiliary pool and additionally use English data (\textsc{+En}), target-language data (\textsc{+Tgt}), bilingual monolingual data (\textsc{+BiMono}), or our proposed code-switched data (\textsc{+Ours}).
English rows are shown in gray to indicate source-language evaluation, which is reported as a retention check rather than the primary target-language transfer metric.
The asterisk (*) indicates that the benchmark is translated into the respective language.
The best performance within each model is highlighted in \textbf{bold}.
}
\label{tab:transposed-domain-ablation-v2}
\end{table*}

%% file: latex/tables/cyber_full.tex
\begin{table*}[!t]
\centering
\setlength{\tabcolsep}{4.5pt}
\definecolor{pointgreen}{HTML}{03C75A}

\resizebox{\textwidth}{!}{
\begin{tabular}{@{} l cccccc | cccccc @{}}
\toprule
& \multicolumn{12}{c}{\textbf{SmolLM3-3B}} \\
\cmidrule(lr){2-13}
& \multicolumn{6}{c|}{\textbf{Korean}} & \multicolumn{6}{c}{\textbf{Swahili}} \\
\cmidrule(lr){2-7} \cmidrule(lr){8-13}
\textbf{Bench.} & Base & RLVR-Base & +En & +Tgt & +BiMono & +Ours & Base & RLVR-Base & +En & +Tgt & +BiMono & +Ours \\
\midrule

\multicolumn{13}{@{}l}{\textbf{Target Language}} \\
\midrule
SecEval* & 40.9 & 44.3 & 50.1 & 49.8 & 51.1 & \cellcolor{pointgreen!15}\textbf{57.3} & 26.9 & 25.8 & 26.8 & \textbf{28.6} & 25.9 & \cellcolor{pointgreen!15}28.3 \\
CyberMetric* & 63.6 & 65.8 & 70.6 & \textbf{80.2} & 72.4 & \cellcolor{pointgreen!15}80.0 & 53.0 & 54.0 & 53.9 & 54.2 & 54.5 & \cellcolor{pointgreen!15}\textbf{55.0} \\
CTI-MCQ* & 33.7 & 40.6 & 43.2 & 45.5 & 43.8 & \cellcolor{pointgreen!15}\textbf{47.1} & 41.4 & 41.9 & 42.1 & 43.2 & 43.4 & \cellcolor{pointgreen!15}\textbf{45.0} \\
\cmidrule(lr){1-13}
\textit{Avg (Tgt)} & 46.07 & 50.23 & 54.63 & 58.50 & 55.77 & \cellcolor{pointgreen!25}\textbf{61.47} & 40.43 & 40.57 & 40.93 & 42.00 & 41.27 & \cellcolor{pointgreen!25}\textbf{42.77} \\
\midrule

\multicolumn{13}{@{}l}{\color{gray} \textbf{English}} \\
\midrule
\color{gray} SecEval & \color{gray} 55.6 & \color{gray} 56.7 & \color{gray} 59.9 & \color{gray} 61.7 & \color{gray} 58.9 & \color{gray} \textbf{62.7} & \color{gray} 55.6 & \color{gray} 56.7 & \color{gray} 59.9 & \color{gray} 65.6 & \color{gray} 65.3 & \color{gray} \textbf{66.6} \\
\color{gray} CyberMetric & \color{gray} 77.0 & \color{gray} 83.0 & \color{gray} 83.6 & \color{gray} \textbf{85.6} & \color{gray} 84.4 & \color{gray} \textbf{85.6} & \color{gray} 77.0 & \color{gray} 83.0 & \color{gray} 83.6 & \color{gray} 86.3 & \color{gray} 86.0 & \color{gray} \textbf{86.6} \\
\color{gray} CTI-MCQ & \color{gray} 50.7 & \color{gray} 54.9 & \color{gray} \textbf{57.5} & \color{gray} 56.2 & \color{gray} 57.1 & \color{gray} 56.5 & \color{gray} 50.7 & \color{gray} 54.9 & \color{gray} 57.5 & \color{gray} \textbf{61.1} & \color{gray} 60.9 & \color{gray} 60.3 \\
\cmidrule(lr){1-13}
\color{gray} \textit{Avg (En)} & \color{gray} 61.10 & \color{gray} 64.87 & \color{gray} 67.00 & \color{gray} 67.83 & \color{gray} 66.80 & \color{gray} \textbf{68.27} & \color{gray} 61.10 & \color{gray} 64.87 & \color{gray} 67.00 & \color{gray} 71.00 & \color{gray} 70.73 & \color{gray} \textbf{71.17} \\
\bottomrule
\end{tabular}
}
\caption{
Comprehensive performance on cybersecurity benchmarks across RLVR data configurations.
\textbf{Target Language} indicates Korean for the left section and Swahili for the right section.
\textsc{Base} denotes the original instruction-tuned model without RLVR.
\textsc{RLVR-Base} denotes RLVR training using only the auxiliary data pool.
The remaining variants include the same auxiliary pool and additionally use English data (\textsc{+En}), target-language data (\textsc{+Tgt}), bilingual monolingual data (\textsc{+BiMono}), or our proposed code-switched data (\textsc{+Ours}).
English rows are shown in gray to indicate source-language evaluation, which is reported as a retention check rather than the primary target-language transfer metric.
The asterisk (*) indicates that the benchmark is translated into the respective language.
The best performance within each model is highlighted in \textbf{bold}.
}
\label{tab:main-domain-ko-sw}
\end{table*}

%% file: latex/tables/component_ablation.tex
\begin{table*}[!t]
\centering
\small
\setlength{\tabcolsep}{6pt}
\definecolor{pointgreen}{HTML}{03C75A}
\begin{tabular*}{\textwidth}{@{\extracolsep{\fill}} l ccccccc}
\toprule
& \multicolumn{7}{c}{\textbf{SmolLM3-3B (Korean)}} \\
\cmidrule(lr){2-8}
\textbf{Bench.} & RLVR-Base & +En & +Tgt & +BiMono & +Opt & +Pass & +Ours \\
\midrule
\multicolumn{8}{@{}l}{\textbf{Target Language}} \\
\midrule
SecEval*     & 44.3 & 50.1 & 49.8          & 51.1 & 52.0 & 54.7 & \cellcolor{pointgreen!15}\textbf{57.3} \\
CyberMetric* & 65.8 & 70.6 & \textbf{80.2} & 72.4 & 71.0 & 77.4 & \cellcolor{pointgreen!15}80.0 \\
CTI-MCQ*     & 40.6 & 43.2 & 45.5          & 43.8 & 45.9 & 46.0 & \cellcolor{pointgreen!15}\textbf{47.1} \\
\cmidrule(lr){1-8}
\textit{Avg (Tgt)} & 50.23 & 54.63 & 58.50 & 55.77 & 56.30 & 59.37 & \cellcolor{pointgreen!25}\textbf{61.47} \\
\midrule
\multicolumn{8}{@{}l}{\color{gray} \textbf{English}} \\
\midrule
\color{gray} SecEval     & \color{gray} 56.7 & \color{gray} 59.9 & \color{gray} 61.7 & \color{gray} 58.9 & \color{gray} 60.3 & \color{gray} 61.2 & \color{gray} \textbf{62.7} \\
\color{gray} CyberMetric & \color{gray} 83.0 & \color{gray} 83.6 & \color{gray} \textbf{85.6} & \color{gray} 84.4 & \color{gray} 84.0 & \color{gray} 84.0 & \color{gray} \textbf{85.6} \\
\color{gray} CTI-MCQ     & \color{gray} 54.9 & \color{gray} \textbf{57.5} & \color{gray} 56.2 & \color{gray} 57.1 & \color{gray} 57.3 & \color{gray} 56.0 & \color{gray} 56.5 \\
\cmidrule(lr){1-8}
\color{gray} \textit{Avg (En)} & \color{gray} 64.87 & \color{gray} 67.00 & \color{gray} 67.83 & \color{gray} 66.80 & \color{gray} 67.20 & \color{gray} 67.07 & \color{gray} \textbf{68.27} \\
\bottomrule
\end{tabular*}
\caption{Component ablation on Korean cybersecurity benchmarks with
SmolLM3-3B. \textsc{Opt-only} and \textsc{Pass-only} extend
\textsc{Bi-Mono} with option-side and passage-level code-switching,
respectively, while \textsc{Ours} combines both components. All variants
follow the same experimental setup. Asterisks denote target-language versions
of the benchmarks.}
\label{tab:component-ablation-main}
\end{table*}

%% file: latex/tables/consistency_table_perbench.tex
\begin{table}[t]
\centering
\small
\begin{tabular}{lccc}
\toprule
(English, Korean) & ($\checkmark$,$\checkmark$) $\uparrow$ & ($\checkmark$,$\times$) $\downarrow$ & ($\times$,$\checkmark$) $\downarrow$ \\
\midrule
\multicolumn{4}{l}{\textit{SecEval}} \\
\midrule
Base & 8.3 & 39.2 & 5.4 \\
En-only & 15.3 & 41.8 & 3.4 \\
Bi-Mono & 16.3 & 41.1 & 3.7 \\
\textbf{Ours} & \textbf{23.8} & \textbf{38.8} & \textbf{2.5} \\
\midrule
\multicolumn{4}{l}{\textit{CyberMetric}} \\
\midrule
Base & 31.4 & 13.8 & 32.0 \\
En-only & 62.0 & 15.6 & 9.6 \\
Bi-Mono & 64.0 & 17.6 & 7.2 \\
\textbf{Ours} & \textbf{74.4} & \textbf{11.2} & \textbf{5.4} \\
\midrule
\multicolumn{4}{l}{\textit{CTI-MCQ}} \\
\midrule
Base & 17.7 & 21.4 & 15.8 \\
En-only & 32.6 & 22.6 & 10.4 \\
Bi-Mono & 33.5 & 22.4 & \textbf{10.0} \\
\textbf{Ours} & \textbf{36.5} & \textbf{20.0} & 10.6 \\
\bottomrule
\end{tabular}
\caption{Detailed cross-lingual consistency comparison for each benchmark. \textbf{Bold} indicates the highest consistency.}
\label{tab:xling_consistency_perbench}
\end{table}

%% file: latex/tables/full_pairwise_judge.tex
\begin{table*}[!t]
\centering\small
\setlength{\tabcolsep}{4pt}
\begin{tabular}{ll rrr c rrr}
\toprule
& & \multicolumn{3}{c}{vs.\ \textsc{En-only}} & & \multicolumn{3}{c}{vs.\ \textsc{Tgt-only}} \\
\cmidrule(lr){3-5}\cmidrule(lr){7-9}
 & Benchmark & Ours\% & En-only\% & tie\% & & Ours\% & Tgt-only\% & tie\% \\
\midrule
\multicolumn{9}{@{}l}{\textbf{Medical}} \\
\cmidrule(lr){1-9}
\multirow{5}{*}{Qwen2.5-3B}
 & MedQA              & \textbf{47.0} & 36.0 & 1.0 && \textbf{48.0} & 41.0 & 0.0 \\
 & MedMCQA            & \textbf{50.0} & 40.0 & 3.0 && 46.0 & \textbf{47.0} & 1.0 \\
 & KMMLU-Pro-Medical  & \textbf{50.0} & 37.0 & 0.0 && \textbf{54.0} & 36.0 & 0.0 \\
 & KorMedMCQA-Doctor  & 46.0 & \textbf{47.0} & 0.0 && 42.0 & \textbf{46.0} & 2.0 \\
 \cmidrule(lr){2-9}
 & \multicolumn{1}{r}{\textit{Average}}
                      & \textit{48.3} & \textit{40.0} & \textit{1.0} && \textit{47.5} & \textit{42.5} & \textit{0.8} \\
\midrule
\multirow{5}{*}{Llama-3.2-3B}
 & MedQA              & 44.0 & \textbf{45.0} & 3.0 && \textbf{55.0} & 39.0 & 0.0 \\
 & MedMCQA            & \textbf{41.0} & 33.0 & 5.0 && 30.0 & \textbf{48.0} & 3.0 \\
 & KMMLU-Pro-Medical  & \textbf{48.0} & 43.0 & 1.0 && \textbf{46.0} & 44.0 & 0.0 \\
 & KorMedMCQA-Doctor  & \textbf{55.0} & 34.0 & 0.0 && \textbf{42.0} & 37.0 & 0.0 \\
 \cmidrule(lr){2-9}
 & \multicolumn{1}{r}{\textit{Average}}
                      & \textit{47.0} & \textit{38.8} & \textit{2.3} && \textit{43.3} & \textit{42.0} & \textit{0.8} \\
\midrule
\multicolumn{9}{@{}l}{\textbf{Cyber}} \\
\cmidrule(lr){1-9}
\multirow{4}{*}{SmolLM3-3B}
 & CTI-MCQ     & \textbf{45.0} & 35.0 &  5.0 && \textbf{69.0} & 26.0 &  0.0 \\
 & CyberMetric & \textbf{31.0} & 30.0 & 13.0 && 25.0 & \textbf{27.0} & 17.0 \\
 & SecEval     & \textbf{33.0} & 30.0 & 12.0 && \textbf{82.0} & 12.0 &  0.0 \\
 \cmidrule(lr){2-9}
 & \multicolumn{1}{r}{\textit{Average}}
                      & \textit{36.3} & \textit{31.7} & \textit{10.0} && \textit{58.7} & \textit{21.7} & \textit{5.7} \\
\bottomrule
\end{tabular}
\caption{Detailed pairwise judge win rate results using GPT-5.5. The target language is Korean. 100 samples are evaluated per benchmark. \textbf{Bold} indicates a higher win rate.}
\label{tab:judge_full_appendix}
\end{table*}

%% file: latex/figures/qual_med_eg_1.tex
\begin{figure*}[t]
\centering

\tcbset{
  variantbox/.style={
    colback=blue!3,       
    colframe=blue!60,     
    boxrule=1pt,          
    arc=3pt,              
    left=6pt, right=6pt, top=4pt, bottom=4pt,
    fonttitle=\bfseries,  
    width=\linewidth
  }
}

\begin{tcolorbox}[variantbox, title=Variant: En-only]
\small
\textbf{\textit{Passage:}}\\
Anti-programmed death-1 (anti-PD-1) and programmed death-ligand 1 (anti-PD-L1) monoclonal antibodies are approved treatments for several malignancies, including nivolumab and pembrolizumab for melanoma and non-small-cell lung cancer (NSCLC), nivolumab for renal cell carcinoma and Hodgkin’s lymphoma, atezolizumab for bladder cancer, and the combination of nivolumab plus ipilimumab for melanoma. While these agents are generally associated with mild toxicity, they can provoke serious immune-related adverse events. \textcolor{red}{\textbf{[MASK]}}. Drug-induced pneumonia may also arise from conventional chemotherapy agents such as docetaxel, gemcitabine, and bleomycin, from targeted therapies like epidermal growth factor receptor inhibitors and mammalian target of rapamycin inhibitors, as well as from radiotherapy.

\vspace{4pt}
\textbf{\textit{Options:}}
\begin{itemize}[leftmargin=1.8em, itemsep=2pt, topsep=0pt, parsep=0pt]
\item[A)] Pneumonia—defined as focal or diffuse inflammation of the lung parenchyma—is one such adverse event; its reported incidence in clinical trials of anti-PD-1/PD-L1 antibodies ranges from 0\,\% to 10\,\% $\checkmark$
\item[B)] Colitis — characterized by ulcerative lesions of the intestinal mucosa — another potential toxicity ; its observed frequency in phase III studies of PD-1/PD-L1 blockers varies between 2\,\% and 12\,\% in treated patients 
\item[C)] Pancreatitis involving inflammatory swelling of the exocrine gland is a rare metabolic side effect ; its noted proportion within large anti-PD-1/PD-L treatment groups hovers around three percent through eight percent across cycles 
\item[D)] Hypothyroidism — characterized by reduced thyroid hormone production and fatigue — a common endocrine immune-related adverse effect ; its occurrence rate across anti-PD-1/PD-L1 studies lies near 15\,\% to 30\,\% globally in adult populations 
\item[E)] Myocarditis — manifested as inflammatory infiltration of cardiac muscle — an infrequent yet severe checkpoint inhibitor toxicity ; its prevalence reported in extensive PD-1/PD-L1 cohorts fluctuates between 1\,\% and 5\,\% overall in patients 
\item[F)] Hepatitis — identified by elevated transaminases and hepatic inflammation — a documented immune-related complication ; its cumulative incidence across pivotal PD-1/PD-L1 trials spans roughly from 3\,\% up to 9\,\% among study participants 
\item[G)] Dermatitis — described as erythematous rash covering variable body surface area — a frequent cutaneous manifestation ; its incidence reported in major anti-PD-01/PD-L1 investigations spans roughly from 20\,\% to 45\,\%, depending on regimen 
\item[H)] Nephritis — presenting with proteinuria haematuria and impaired renal function — another recognized renal toxicity ; its documented frequency among PD-1/PD-L1 recipients ranges approximately from 2\,\% up to 8\,\% during therapy in clinical trials 
\item[I)] Encephalitis — defined as brain parenchymal inflammation with neurological symptoms — an uncommon neurotoxic event ; its estimated rate among pooled PD-1/PD-L1 patient data sets runs from five per cent up to fifteen per cent during follow up 
\end{itemize}
\end{tcolorbox}

\caption{Qualitative example — Medical (Part 1/2). Same source passage shown across 3 variants. \textcolor{red}{[MASK]} marks the masked answer span; the ground-truth option ($\checkmark$) is what fills it.}
\label{fig:qual-medical-1}
\end{figure*}

%% file: latex/figures/qual_med_eg_2.tex
\begin{figure*}[t]
\centering

\tcbset{
  variantbox/.style={
    colback=blue!3,       
    colframe=blue!60,     
    boxrule=1pt,          
    arc=3pt,              
    left=6pt, right=6pt, top=4pt, bottom=4pt,
    fonttitle=\bfseries,  
    width=\linewidth
  }
}

\begin{tcolorbox}[variantbox, title=Variant: Ko-only]
\small
\textbf{\textit{Passage:}}\\
Anti-programmed death-1 (anti-PD-1) 및 programmed death-ligand 1 (anti-PD-L1) 단클론 항체는 흑색종 및 비소세포폐암(NSCLC)에 대한 nivolumab 및 pembrolizumab, 신세포암 및 호지킨 림프종에 대한 nivolumab, 방광암에 대한 atezolizumab, 그리고 흑색종에 대한 nivolumab과 ipilimumab 병용 요법을 포함하여 여러 악성 종양에 대해 승인된 치료제입니다. 이러한 약제들은 일반적으로 경미한 독성과 관련이 있지만, 심각한 면역 관련 이상 반응을 유발할 수 있습니다. \textcolor{red}{\textbf{[MASK]}}. 약물 유발 폐렴은 docetaxel, gemcitabine, bleomycin과 같은 전통적인 화학요법제, epidermal growth factor receptor inhibitors 및 mammalian target of rapamycin inhibitors와 같은 표적 치료제, 그리고 방사선 치료로 인해 발생할 수도 있습니다. 

\vspace{4pt}
\textbf{\textit{Options:}}
\begin{itemize}[leftmargin=1.8em, itemsep=2pt, topsep=0pt, parsep=0pt]
\item[A)] 폐렴—폐 실질의 국소적 또는 미만성 염증으로 정의됨—은 그러한 이상 반응 중 하나입니다; anti-PD-1/PD-L1 항체의 임상 시험에서 보고된 발생률은 0\,\%에서 10\,\% 사이입니다. $\checkmark$
\item[B)] 장 점막의 궤양성 병변을 특징으로 하는 결장염은 또 다른 잠재적 독성입니다. PD-1/PD-L1 차단제의 임상 3상 연구에서 관찰된 치료 환자에서의 발생 빈도는 2\,\%에서 12\,\% 사이로 다양합니다. 
\item[C)] 외분비선의 염증성 부종을 동반하는 췌장염은 드문 대사성 부작용입니다; 대규모 anti-PD-1/PD-L1 치료군 내에서 관찰되는 비율은 사이클 전반에 걸쳐 3\,\%에서 8\,\% 사이를 오갑니다. 
\item[D)] 갑상선 기능 저하증 — 갑상선 호르몬 생성 감소와 피로를 특징으로 함 — 흔한 내분비 면역 관련 이상 반응 ; 전 세계 성인 인구 대상 anti-PD-1/PD-L1 연구에서의 발생률은 15\,\%에서 30\,\% 사이입니다. 
\item[E)] 심근염 — 심근의 염증성 침윤으로 나타나는 — 드물지만 심각한 checkpoint inhibitor 독성; 광범위한 PD-1/PD-L1 코호트에서 보고된 유병률은 환자 전체에서 1\,\%에서 5\,\% 사이에서 변동합니다. 
\item[F)] 간염 — 상승된 트랜스아미나제와 간 염증으로 식별됨 — 기록된 면역 관련 합병증; 주요 PD-1/PD-L1 임상 시험 전반에 걸친 연구 참여자들의 누적 발생률은 대략 3\,\%에서 9\,\%에 이릅니다. 
\item[G)] 피부염 — 다양한 체표면적을 덮는 홍반성 발진으로 설명됨 — 빈번한 피부 증상; 주요 anti-PD-1/PD-L1 연구에서 보고된 발생률은 요법에 따라 대략 20\,\%에서 45\,\%에 이릅니다. 
\item[H)] 신염 — 단백뇨, 혈뇨 및 신기능 저하를 동반함 — 또 다른 알려진 신독성; 임상 시험 중 PD-1/PD-L1 투여군에서 보고된 빈도는 치료 기간 동안 약 2\,\%에서 최대 8\,\%에 이릅니다. 
\item[I)] 뇌염 — 신경학적 증상을 동반한 뇌 실질의 염증으로 정의됨 — 은 흔하지 않은 신경독성 사건입니다; 통합된 PD-1/PD-L1 환자 데이터 세트에서 추정되는 발생률은 추적 관찰 기간 동안 5\,\%에서 15\,\% 사이입니다. 
\end{itemize}
\end{tcolorbox}

\vspace{6pt} %

\begin{tcolorbox}[variantbox, title=Variant: Ko$\to$En (Passage-level Code-switching)]
\small
\textbf{\textit{Passage:}}\\
Anti-programmed death-1 (anti-PD-1) 및 programmed death-ligand 1 (anti-PD-L1) monoclonal antibodies는 melanoma 및 non-small cell lung cancer(NSCLC)에 대한 nivolumab 및 pembrolizumab, renal cell carcinoma 및 Hodgkin lymphoma에 대한 nivolumab, bladder cancer에 대한 atezolizumab, 그리고 melanoma에 대한 nivolumab과 ipilimumab combination therapy를 포함하여 various malignant tumors에 대해 approved therapeutics입니다. 이러한 약제들은 generally mild toxicity와 관련이 있지만, 심각한 immune-related adverse reactions을 유발할 수 있습니다. \textcolor{red}{\textbf{[MASK]}}. 약물 유발 폐렴은 docetaxel, gemcitabine, bleomycin과 같은 전통적인 화학요법제, epidermal growth factor receptor inhibitors 및 mammalian target of rapamycin inhibitors와 같은 표적 치료제, 그리고 방사선 치료로 인해 발생할 수도 있습니다.

\vspace{4pt}
\textbf{\textit{Options:}}
\begin{itemize}[leftmargin=1.8em, itemsep=2pt, topsep=0pt, parsep=0pt]
\item[A)] 폐렴—폐 실질의 국소적 또는 미만성 염증으로 정의됨—은 그러한 이상 반응 중 하나입니다; anti-PD-1/PD-L1 항체의 임상 시험에서 보고된 발생률은 0\,\%에서 10\,\% 사이입니다. $\checkmark$
\item[B)] 장 점막의 궤양성 병변을 특징으로 하는 결장염은 또 다른 잠재적 독성입니다. PD-1/PD-L1 차단제의 임상 3상 연구에서 관찰된 치료 환자에서의 발생 빈도는 2\,\%에서 12\,\% 사이로 다양합니다. 
\item[C)] 외분비선의 염증성 부종을 동반하는 췌장염은 드문 대사성 부작용입니다; 대규모 anti-PD-1/PD-L1 치료군 내에서 관찰되는 비율은 사이클 전반에 걸쳐 3\,\%에서 8\,\% 사이를 오갑니다. 
\item[D)] 갑상선 기능 저하증 — 갑상선 호르몬 생성 감소와 피로를 특징으로 함 — 흔한 내분비 면역 관련 이상 반응 ; 전 세계 성인 인구 대상 anti-PD-1/PD-L1 연구에서의 발생률은 15\,\%에서 30\,\% 사이입니다. 
\item[E)] 심근염 — 심근의 염증성 침윤으로 나타나는 — 드물지만 심각한 checkpoint inhibitor 독성; 광범위한 PD-1/PD-L1 코호트에서 보고된 유병률은 환자 전체에서 1\,\%에서 5\,\% 사이에서 변동합니다. 
\item[F)] 간염 — 상승된 트랜스아미나제와 간 염증으로 식별됨 — 기록된 면역 관련 합병증; 주요 PD-1/PD-L1 임상 시험 전반에 걸친 연구 참여자들의 누적 발생률은 대략 3\,\%에서 9\,\%에 이릅니다. 
\item[G)] 피부염 — 다양한 체표면적을 덮는 홍반성 발진으로 설명됨 — 빈번한 피부 증상; 주요 anti-PD-1/PD-L1 연구에서 보고된 발생률은 요법에 따라 대략 20\,\%에서 45\,\%에 이릅니다. 
\item[H)] 신염 — 단백뇨, 혈뇨 및 신기능 저하를 동반함 — 또 다른 알려진 신독성; 임상 시험 중 PD-1/PD-L1 투여군에서 보고된 빈도는 치료 기간 동안 약 2\,\%에서 최대 8\,\%에 이릅니다. 
\item[I)] 뇌염 — 신경학적 증상을 동반한 뇌 실질의 염증으로 정의됨 — 은 흔하지 않은 신경독성 사건입니다; 통합된 PD-1/PD-L1 환자 데이터 세트에서 추정되는 발생률은 추적 관찰 기간 동안 5\,\%에서 15\,\% 사이입니다. 
\end{itemize}
\end{tcolorbox}

\caption{Qualitative example — Medical (Part 2/2). Continuation of the 3 variants.}
\label{fig:qual-medical-2}
\end{figure*}

%% file: latex/figures/qual_cyber_eg_1.tex
\begin{figure*}[t]
\centering

\tcbset{
  variantbox/.style={
    colback=blue!3,       
    colframe=blue!60,     
    boxrule=1pt,          
    arc=3pt,              
    left=6pt, right=6pt, top=4pt, bottom=4pt,
    fonttitle=\bfseries,  
    width=\linewidth
  }
}

\begin{tcolorbox}[variantbox, title=Variant: En-only]
\small
\textbf{\textit{Passage:}}\\
Sender Policy Framework (SPF) is an email authentication system used to verify that a message originates from an authorized server for the sender’s domain. \textcolor{red}{\textbf{[MASK]}} By enforcing this check, SPF prevents attackers from forging (spoofing) the “From” address of a legitimate domain, which is a common technique for phishing and spam campaigns. 

\vspace{4pt}
\textbf{\textit{Options:}}
\begin{itemize}[leftmargin=1.8em, itemsep=2pt, topsep=0pt, parsep=0pt]
\item[A)] When configuring SPF, administrators must add a CNAME record pointing to an external validation service, which then returns a list of allowed sending hosts during each SMTP handshake. 
\item[B)] Activating SPF requires adding a DNS A record for each permitted mail server, after which receiving systems perform a reverse‑DNS lookup to match the connecting IP to the A records. 
\item[C)] Deploying SPF for a domain creates a DNS TXT record that lists the IP addresses of mail servers permitted to send mail on behalf of that domain. Once the record propagates to DNS resolvers worldwide, each receiving mail server can query the sender’s DNS zone during the SMTP transaction. The receiving server checks whether the connecting IP address appears in the SPF record; if it does, the message is accepted and delivered, otherwise it is rejected or marked as suspicious. $\checkmark$
\item[D)] The SPF setup process creates a DNS TXT entry that enumerates permitted sending hosts; receiving mail servers then validate the connecting client’s IP against this entry during SMTP. 
\item[E)] Setting up SPF entails publishing a DNS TXT record containing the authorized IP blocks, after which each inbound mail server performs an SPF‑PASS lookup before delivering the message. 
\item[F)] Implementing SPF involves publishing a DNS MX record that enumerates all mail exchangers, and the receiver validates the HELO/EHLO hostname against this list before accepting mail. 
\item[G)] Configuring SPF means inserting a DNS TXT entry with a "v=spf1" tag followed by “include” mechanisms, and receivers compare the sender’s DKIM signature against this policy. 
\item[H)] To set up SPF, you create a DNS SRV record containing the authorized IP ranges, and mail servers query this record on every inbound connection to decide acceptance. 
\item[I)] During SPF deployment, a DNS TXT record is published that lists authorized IPs, and receiving servers use this list to compute a hash of the sender’s envelope‑from domain for verification. 
\end{itemize}
\end{tcolorbox}

\vspace{6pt} %

\begin{tcolorbox}[variantbox, title=Variant: Ko-only]
\small
\textbf{\textit{Passage:}}\\
Sender Policy Framework (SPF)는 메시지가 발신자 도메인에 대해 권한이 있는 서버에서 생성되었는지 확인하는 데 사용되는 이메일 인증 시스템입니다. \textcolor{red}{\textbf{[MASK]}} 이 검사를 강제함으로써, SPF는 공격자가 피싱 및 스팸 캠페인에서 흔히 사용하는 기술인 합법적인 도메인의 “From” 주소를 위조(spoofing)하는 것을 방지합니다. 

\vspace{4pt}
\textbf{\textit{Options:}}
\begin{itemize}[leftmargin=1.8em, itemsep=2pt, topsep=0pt, parsep=0pt]
\item[A)] SPF를 구성할 때, 관리자는 외부 검증 서비스로 가리키는 CNAME 레코드를 추가해야 하며, 해당 서비스는 각 SMTP 핸드쉐이크 동안 허용된 발신 호스트 목록을 반환합니다. 
\item[B)] SPF를 활성화하려면 허용된 각 메일 서버에 대해 DNS A 레코드를 추가해야 하며, 이후 수신 시스템은 연결된 IP와 A 레코드를 매칭하기 위해 역 DNS 조회를 수행합니다. 
\item[C)] 도메인에 SPF를 배포하면 해당 도메인을 대신하여 메일을 보낼 수 있는 메일 서버의 IP 주소를 나열한 DNS TXT 레코드가 생성됩니다. 레코드가 전 세계 DNS 리졸버에 전파되면, 각 수신 메일 서버는 SMTP 트랜잭션 중에 발신자의 DNS 영역을 조회할 수 있습니다. 수신 서버는 연결된 IP 주소가 SPF 레코드에 포함되어 있는지 확인하고, 포함되어 있으면 메시지를 수락하고 전달하며, 그렇지 않으면 거부하거나 의심스러운 것으로 표시합니다. $\checkmark$
\item[D)] SPF 설정 과정에서는 허용된 발신 호스트를 열거한 DNS TXT 엔트리를 생성하고, 수신 메일 서버는 SMTP 중에 연결된 클라이언트의 IP를 이 엔트리와 대조하여 검증합니다. 
\item[E)] SPF를 설정하려면 허가된 IP 블록을 포함한 DNS TXT 레코드를 게시하고, 이후 각 인바운드 메일 서버는 메시지를 전달하기 전에 SPF‑PASS 조회를 수행합니다. 
\item[F)] SPF를 구현하려면 모든 메일 교환기를 열거한 DNS MX 레코드를 게시하고, 수신자는 메일을 수락하기 전에 HELO/EHLO 호스트명을 이 목록과 대조하여 검증합니다. 
\item[G)] SPF를 구성한다는 것은 "v=spf1" 태그와 뒤에 이어지는 “include” 메커니즘을 포함한 DNS TXT 엔트리를 삽입하는 것을 의미하며, 수신자는 발신자의 DKIM 서명을 이 정책과 비교합니다. 
\item[H)] SPF를 설정하려면 허가된 IP 범위를 포함한 DNS SRV 레코드를 생성하고, 메일 서버는 각 인바운드 연결 시 이 레코드를 조회하여 수락 여부를 결정합니다. 
\item[I)] SPF 배포 중에 허가된 IP를 나열한 DNS TXT 레코드가 게시되며, 수신 서버는 이 목록을 사용해 발신자의 envelope‑from 도메인의 해시를 계산하여 검증합니다. 
\end{itemize}
\end{tcolorbox}

\caption{Qualitative example — Cybersecurity (Part 1/2). Same source passage shown across 4 variants. \textcolor{red}{[MASK]} marks the masked answer span; the ground-truth option ($\checkmark$) is what fills it.}
\label{fig:qual-cybersecurity-1}
\end{figure*}

%% file: latex/figures/qual_cyber_eg_2.tex
\begin{figure*}[t]
\centering

\tcbset{
  variantbox/.style={
    colback=blue!3,       
    colframe=blue!60,     
    boxrule=1pt,          
    arc=3pt,              
    left=6pt, right=6pt, top=4pt, bottom=4pt,
    fonttitle=\bfseries,  
    width=\linewidth
  }
}

\begin{tcolorbox}[variantbox, title=Variant: SW-only]
\small
\textbf{\textit{Passage:}}\\
Sender Policy Framework (SPF) ni mfumo wa uthibitishaji wa barua pepe unaotumiwa kuhakiki kuwa ujumbe unatokana na seva iliyoidhinishwa kwa ajili ya domain ya mtumaji. \textcolor{red}{\textbf{[MASK]}} Kwa kusimamia ukaguzi huu, SPF huzuia washambuliaji kutengeneza (spoofing) anwani ya "From" ya domain halali, ambayo ni mbinu ya kawaida kwa kampeni za phishing na spam. 

\vspace{4pt}
\textbf{\textit{Options:}}
\begin{itemize}[leftmargin=1.8em, itemsep=2pt, topsep=0pt, parsep=0pt]
\item[A)] Wakati wa kusanidi SPF, wasimamizi lazima waongeze rekodi ya CNAME inayoelekeza kwenye huduma ya nje ya uhakiki, ambayo kisha hurudisha orodha ya wenyeji wanaoruhusiwa kutuma wakati wa kila SMTP handshake. 
\item[B)] Kuamsha SPF kunahitaji kuongeza rekodi ya DNS A kwa kila seva ya barua pepe iliyoruhusiwa, baada ya hapo mifumo inayopokea hufanya reverse-DNS lookup ili kuoanisha IP inayojiunganisha na rekodi za A. 
\item[C)] Kuweka SPF kwa ajili ya domain kunatengeneza rekodi ya DNS TXT inayoorodhesha IP addresses za mail servers zinazoruhusiwa kutuma barua pepe kwa niaba ya domain hiyo. Mara tu rekodi inapoelea kwenye DNS resolvers duniani kote, kila seva ya barua pepe inayopokea inaweza kuulizia DNS zone ya mtumaji wakati wa muamala wa SMTP. Seva inayopokea hukagua ikiwa anwani ya IP inayojiunganisha inaonekana katika rekodi ya SPF; ikiwa inaonekana, ujumbe unakubaliwa na kuwasilishwa, vinginevyo unakataliwa au kuwekwa alama kama wa shaka. $\checkmark$
\item[D)] Mchakato wa usanidi wa SPF unaunda ingizo la DNS TXT ambalo linaorodhesha host zinazoruhusiwa kutuma; seva za kupokea barua pepe kisha zinathibitisha IP ya mteja anayeunganishwa dhidi ya ingizo hili wakati wa SMTP. 
\item[E)] Kuandaa SPF kunahusisha kuchapisha rekodi ya DNS TXT inayojumuisha vizuizi vya IP vilivyoidhinishwa, baada ya hapo kila seva ya barua pepe inayopokea inafanya utafutaji wa SPF‑PASS kabla ya kuwasilisha ujumbe. 
\item[F)] Kutekeleza SPF kunahusisha kuchapisha rekodi ya DNS MX inayoorodhesha mail exchangers zote, na mpokeaji huhakiki hostname ya HELO/EHLO dhidi ya orodha hii kabla ya kukubali barua pepe. 
\item[G)] Kusanidi SPF inamaanisha kuingiza rekodi ya DNS TXT yenye tag ya "v=spf1" ikifuatiwa na mifumo ya “include”, na wapokeaji hulinganisha saini ya DKIM ya mtumaji dhidi ya sera hii. 
\item[H)] Ili kuweka SPF, unatengeneza rekodi ya DNS SRV inayojumuisha masafa ya IP yaliyoidhinishwa, na seva za barua pepe huulizia rekodi hii kwenye kila muunganisho unaoingia ili kuamua kukubalika. 
\item[I)] Wakati wa utekelezaji wa SPF, rekodi ya DNS TXT huchapishwa ambayo inaorodhesha IP zilizoidhinishwa, na seva zinazopokea hutumia orodha hii kukokotoa hash ya envelope‑from domain ya mtumaji kwa ajili ya uhakiki. 
\end{itemize}
\end{tcolorbox}

\vspace{6pt}

\begin{tcolorbox}[variantbox, title=Variant: EN$\to$SW (Passage-level Code-switching)]
\small
\textbf{\textit{Passage:}}\\
Sender Policy Framework (SPF) ni mfumo wa uthibitishaji wa barua pepe used to verify kwamba ujumbe unatoka kutoka kwa server iliyoidhinishwa kwa ajili ya domain ya mtumaji. \textcolor{red}{\textbf{[MASK]}} Kwa kusimamia ukaguzi huu, SPF inazuia wadukuzi kutoka kughushi (spoofing) anuani ya "From" ya domain halali, ambayo ni mbinu ya kawaida kwa kampeni za phishing na spam. 

\vspace{4pt}
\textbf{\textit{Options:}}
\begin{itemize}[leftmargin=1.8em, itemsep=2pt, topsep=0pt, parsep=0pt]
\item[A)] When configuring SPF, administrators must add a CNAME record pointing to an external validation service, which then returns a list of allowed sending hosts during each SMTP handshake. 
\item[B)] Activating SPF requires adding a DNS A record for each permitted mail server, after which receiving systems perform a reverse‑DNS lookup to match the connecting IP to the A records. 
\item[C)] Deploying SPF for a domain creates a DNS TXT record that lists the IP addresses of mail servers permitted to send mail on behalf of that domain. Once the record propagates to DNS resolvers worldwide, each receiving mail server can query the sender’s DNS zone during the SMTP transaction. The receiving server checks whether the connecting IP address appears in the SPF record; if it does, the message is accepted and delivered, otherwise it is rejected or marked as suspicious. $\checkmark$
\item[D)] The SPF setup process creates a DNS TXT entry that enumerates permitted sending hosts; receiving mail servers then validate the connecting client’s IP against this entry during SMTP. 
\item[E)] Setting up SPF entails publishing a DNS TXT record containing the authorized IP blocks, after which each inbound mail server performs an SPF‑PASS lookup before delivering the message. 
\item[F)] Implementing SPF involves publishing a DNS MX record that enumerates all mail exchangers, and the receiver validates the HELO/EHLO hostname against this list before accepting mail. 
\item[G)] Configuring SPF means inserting a DNS TXT entry with a "v=spf1" tag followed by “include” mechanisms, and receivers compare the sender’s DKIM signature against this policy. 
\item[H)] To set up SPF, you create a DNS SRV record containing the authorized IP ranges, and mail servers query this record on every inbound connection to decide acceptance. 
\item[I)] During SPF deployment, a DNS TXT record is published that lists authorized IPs, and receiving servers use this list to compute a hash of the sender’s envelope‑from domain for verification. 
\end{itemize}
\end{tcolorbox}

\caption{Qualitative example — Cybersecurity (Part 2/2). Continuation of the 4 variants.}
\label{fig:qual-cybersecurity-2}
\end{figure*}